\documentclass[journal]{IEEEtran}

\usepackage{amsfonts,amsmath}
\usepackage{cite}
\usepackage{amsthm}
\usepackage{graphicx}
\usepackage{times} % Add all your packages here
\usepackage[linesnumbered,ruled, vlined]{algorithm2e} % vlined is used to replace end with a vertical line
\usepackage{booktabs}
\usepackage{multirow}
\usepackage{epsfig}
\usepackage{epstopdf}
\usepackage{longtable}
\usepackage{latexsym}
\usepackage{tabularx}
\usepackage{float}
\usepackage{tikz}
\usepackage{pgfplots}
\usepackage{placeins}

\usepackage{caption}
\usepackage{subcaption}
\usepackage{color,soul}
\usepackage{todonotes}
\usepackage{lmodern}
\usepackage{array}
\usepackage{colortbl}
\usepackage[mathcal]{euscript}
\usepackage{titlesec}
\usepackage{jabbrv}
\makeatletter
\def\ps@IEEEtitlepagestyle{%
  \def\@oddhead{\mycopyrightnotice}%
  \def\@evenhead{}%
}
\def\mycopyrightnotice{%
  {\footnotesize {\begin{tabular}[t]{@{}l@{}}This work has been submitted to the IEEE for possible publication. Copyright may be transferred without notice, after which this version may\\ no longer be accessible.\end{tabular}}}% <--- Change here
  \gdef\mycopyrightnotice{}% just in case
}
\makeatother

\makeatletter
\def\Gin@extensions{%
	.png,.pdf,.jpg,.mps,.jpeg,.jbig2,.jb2,%
	.PNG,.PDF,.JPG,.JPEG,.JBIG2,.JB2,%
}%
\makeatother

\SetKwComment{Comment}{/* }{ */}

\graphicspath{{Figs/}}

\makeatletter
\renewcommand{\@algocf@capt@plain}{above}% formerly {bottom}
\newcommand{\cc}{\cellcolor[rgb]{ .92,  .92,  .92}}
\makeatother

\titlespacing*{\section}
{0pt}{1.5ex plus 0.5ex minus .2ex}{1.3ex plus .2ex}
\titlespacing*{\subsection}
{0pt}{1.5ex plus 0.5ex minus .2ex}{1.3ex plus .2ex}

\begin{document}
% {\footnotesize This work has been submitted to the IEEE for possible publication. Copyright may be transferred without notice, after which this version may no longer be accessible.\hfill}

\title{Vector Autoregressive Evolution for Dynamic Multi-Objective Optimisation}

% \title{Vector Autoregression with Environment Aware Hypermutation for Dynamic Multi-Objective Optimisation}

\author{Shouyong~Jiang,~\IEEEmembership{Member,~IEEE},
Yong~Wang,~\IEEEmembership{Senior Member,~IEEE}, 
Yaru~Hu,
Qingyang~Zhang, and
Shengxiang~Yang,~\IEEEmembership{Senior Member,~IEEE}
%\thanks{Manuscript received September 22, 2023; revised \today. This work was supported by the Engineering and Physical Sciences Research Council (EPSRC) of UK under Grant EP/K001310/1.}
% \thanks{Shouyong~Jiang and Yong~Wang are with School of Automation, Central South University, Changsha 410083, China (email: ywang@csu.edu.cn).}
\thanks{Shouyong~Jiang and Yong~Wang are with School of Automation, Central South University, China (email: \{sjiang, ywang\}@csu.edu.cn).}
\thanks{Yaru~Hu is with School of Computer Science and School of Cyberspace Security, Xiangtan University, China (email: huyaru1199@gmail.com).}
\thanks{Qingyang~Zhang is with School of Computer Science and Technology, Jiangsu Normal University, China (email: sweqyian@126.com).}
% \thanks{Shengxiang~Yang is with the Centre for Computational 
%   Intelligence (CCI), School of Computer Science and Informatics, De Montfort 
%   Univesity, The Gateway, Leicester LE1 9BH, U.K. (email: syang@dmu.ac.uk).}
\thanks{Shengxiang~Yang is with School of Computer Science and Informatics,  De Montfort Univesity, U.K. (email: syang@dmu.ac.uk).}

}

\maketitle

\begin{abstract}
Dynamic multi-objective optimisation (DMO) handles optimisation problems with multiple (often conflicting) objectives in varying environments. Such problems pose various challenges to evolutionary algorithms % which have popularly been used to solve complex optimisation problems,
due to their dynamic nature and resource restrictions in changing environments. This paper proposes vector autoregressive evolution (VARE) consisting of vector autoregression (VAR) and environment-aware hypermutation (EAH) to address environmental changes in DMO. VARE builds a VAR model that considers mutual relationship between decision variables to effectively predict the moving solutions in dynamic environments. Additionally, VARE introduces EAH to address the blindness of existing hypermutation strategies in increasing population diversity in dynamic scenarios where predictive approaches are unsuitable. A seamless integration of VAR and EAH in an environment-adaptive manner makes VARE effective to handle a wide range of dynamic environments and competitive with several popular DMO algorithms, as demonstrated in extensive experimental studies. Specially, the proposed algorithm is computationally 50 times faster than two widely-used algorithms (i.e., Tr-RM-MEDA and MOEA/D-SVR) while producing significantly better results. 
\end{abstract}


\IEEEpeerreviewmaketitle

\section{Introduction}
\IEEEPARstart{D}{ynamic} multi-objective optimisation problems (DMOPs) refer to optimisation problems with multiple objectives to be optimised simultaneously in dynamic environments that make problems change over time \cite{Farina2004deb,Jiang2022survey}. Representative dynamics in DMOPs include dynamic constraints, time-dependent fitness landscapes and time-varying number of objectives/variables. The solution to DMOPs is a sequence of Pareto-optimal sets (PS) each representing the tradeoff between objectives at a particular time of changing environments. The mapping of PS from its decision space to objective space is a sequence of Pareto-optimal fronts (PF). DMOPs have been increasingly studied due to their significance in both fundamental and applied research \cite{Eaton2017ACO,Zhang2011AIS,huang2023large,fu2022multiobjective,peng2021multiobjective}.

The multi-objective and dynamic nature of DMOPs distinguishes them from static optimisation problems that have been well documented in the literature, thus requiring effort to develop algorithms to solve them. Evolutionary algorithms (EAs \cite{yu2021dynamic,yang2022local}) have been popularly used to solve DMOPs by employing a population of agents to search for the PS/PF, under Darwin's theory of evolution. After two decades of research, many dynamic multi-objective EAs (DMOEAs) have been developed for dynamic multi-objective optimisation (DMO). These algorithms differ mainly in how changes are handled when detected, and they can be classified into different categories.

The first category includes algorithms that aim to increase/maintain population diversity in dynamic environments, since changes can harm population diversity especially in severely changing environments. A simple approach for diversity increase is the introduction of random solutions or creation of highly mutated solutions to replace some members of the search population right after a change\cite{Deb2007DNSGA2,Jiang2016SGEA}. Such diversity-increase approaches have been widely used to tackle simple environmental changes 
\cite{Ma2021feature}. Some other approaches focus on diversity maintenance over time regardless of changes, including random immigration \cite{Hu2023layered,Zhang2020evo}.
% diversity-enhancing reproduction \cite{Zhang2020evo}, and two-archive maintenance \cite{Chen2017CLY}.
The second category of DMOEAs focuses on the prediction of the changing PS/PF and creates initial population based on prediction results, for each new environment. %Such DMOEAs often assume that changes are predictable or the PSs in a few consecutive environments are similar. 
A number of predictive models have been applied to DMO, including autoregression \cite{Hatzakis2006FL,Zhou2014PPS}, Kalman filter \cite{Muruganantham2016KF,Rambabu2019mixture}, and machine learning \cite{zhang2022solving,Cao2019MOEADSVR,liu2023cooperative}. Recent studies have observed that unidirectional prediction approaches do not work well for DMOPs whose PS segments move in different directions after a change \cite{Rong2018multidirectional}. Thus, multi-directional prediction of PS has gained increasing attention \cite{Ma2021feature,Hu2020solving}. Especially, the recently developed MOEA/D-SVR \cite{Cao2019MOEADSVR} takes the advantage of multiple search directions in MOEA/D \cite{Zhang2007moead} and that of SVR to predict the move of solutions associated with each search direction of MOEA/D. However, MOEA/D-SVR is computationally intensive as it requires to build a SVR regressor for each decision variable, in each search direction.

Multi-population is another category of DMOEAs and has the inherent advantage of maintaining population diversity due to the use of multiple subpopulations targeting different search regions \cite{Goh2009dCOEA,zheng2023dynamic}. %Multi-population is also beneficial when environmental changes do not affect all search regions because only the subpopulations in the affected regions need to make a response to changes. 
%Multi-population approaches usually involve cooperation and competition between subpopulations, and researchers have proposed various techniques to 
Various multi-population approaches have been developed, including dynamic competitive-cooperative EAs (dCOEA) \cite{Goh2009dCOEA}, multi-population particle swarm optimisation \cite{Liu2018pso,chen2019multi}, cooperative co-evolutionary algorithms \cite{Gong2019similarity}, and artificial immune systems \cite{Shang2014quantum}. These algorithms differ mainly in how subpopulations are coordinated to balance cooperation and competition in order to handle changes effectively.

Saving good solutions of historical environments in a memory archive and then retrieving them as necessary is another approach to handle environmental changes. Memory-based studies have concentrated on data management in memory pool, and this includes the storage of special solutions \cite{xu2018memory}, maintenance of memory space \cite{Peng2015novel}, information retrieval \cite{Wang2019SSA,sahmoud2016memory}. 

%This is particularly useful when dealing with changes that are similar to some previous ones, as archived solutions retrieved from a similar environment can be used again. Much effort has been made as to memory maintenance and information retrieval \cite{Goh2009dCOEA}. 

In addition, other kinds of techniques have been proposed for DMO, including reinforcement learning\cite{zou2021reinforcement}, transfer learning\cite{Jiang2017transfer,Jiang2020fast,yan2023inter}, local search \cite{Azzouz2017localsearch}, preference or reference-guided search \cite{Zou2017prediction,Li2019special}, and dynamics-inspired responses \cite{Azzouz2017dynamic,Sahmoud2018type,Ou2019novel,Zhang2019novel}. These techniques have been systematically reviewed in \cite{Jiang2022survey}.

%In what follows, we provide a brief overview of several important change handling approaches. 

% Diversity-based approaches aim to maintain population diversity in dynamic environments. This kind of approaches is motivated by the fact that environmental changes can lead to the deterioration of population diversity. The loss of population diversity, especially when the environment changes severely, is harmful to the search in the new environment. 

In this paper, we focus on popular prediction approaches for DMOPs as we are motivated by the following observations from existing prediction approaches.
% \begin{enumerate}
% \item 

{\it a)} Existing prediction approaches build single-output predictors (albeit often written in vector form) for decision variables under the biased assumption that decision variables are independent from each other. That is, a variable's current value is only correlated with its lagged values from previous time periods. As a result, they fail to capture mutual relationship between variables, i.e., variable dependencies, leading to poor prediction of PS that are determined by all variables collectively. %These approaches are essentially , albeit often written in vector form.
In addition, popular approaches such as centroid or special point based prediction \cite{Zhou2014PPS,Jiang2016SGEA, Jiang2022survey} simply use the same stepsize for their single-output predictors for all variables. However, the amount of change required for distinct variables to reach the new PS is probably different in many scenarios. Recently, MOEA/D-SVR \cite{Cao2019MOEADSVR} has introduced separate predictive models for different variable, overcoming the stepsize issue, but it does not consider variable dependencies and suffers from intensive model building especially when tackling DMOPs with a large number of variables. A straightforward idea to address the above issue is to build multi-output predictors that can predict values for multiple variables all at once, but multi-output predictive models require large data to train especially when a large number of output variables is needed, which is impractical for DMO. Here, we propose vector autoregressive modelling in conduction with dimensionality reduction to overcome this challenge. 

{\it b)} On the other hand, prediction approaches may be challenged in some situations, such as dynamic environments that does not exhibit regular patterns \cite{Jiang2019SDP} and early stage of search that has not collected enough data to build predictors; therefore, other change response approaches should be used to complement prediction approaches in order to handle a wider range of (potentially unpredictable) dynamic features. The choice of hypermutation rather than random immigration is grounded by the fact that the former has a tunable parameter, i.e., mutation distribution index $\eta$, making it more controllable than the latter in increasing population diversity. However, current hypermutation studies use a predefined $\eta$ to create mutated solutions from existing ones. However, there is no reason for $\eta$ to be preset to same values for different problems and environments.
% \end{enumerate}

We therefore argue that adaptive $\eta$ in concert with environment changes be preferred to create environment-aware hypermutation for maintaining population diversity when prediction approaches are unsuitable for dynamic environments in question. The main contributions of this paper are summarised as follows:
\begin{itemize}
\item A vector autoregressive model that considers mutual relationship between the decision variables of candidate solutions is proposed for population prediction to tackle environmental changes. Dimensionality reduction is applied to address dense parameterisation of such models for DMO that often involves a high number of decision variables.

\item In view of the blindness of current hypermutation in diversity increase for dynamic environments, environment-aware hypermutation (EAH) is proposed to rationally adjust population diversity dynamically. 

\item An environment-adaptive integration of the above two strategies is proposed, allowing smart choice of strategies in response to environmental changes.

\item The proposed algorithm is computationally fast, requiring only 2\% runtime of popular algorithms like Tr-RM-MEDA \cite{Jiang2017transfer} and MOEA/D-SVR \cite{Cao2019MOEADSVR} while producing competitive results.
\end{itemize}

The rest of this paper is organised as follows. Section II presents the proposed algorithm for DMO in detail. Section III presents experimental environments and settings, followed by results and analysis in Section IV. Section V discusses key components of the proposed algorithm and parameter sensitivity. Section VI concludes this paper and hints future investigation.

\section{VARE: The Proposed Approach for DMO}
The proposed VARE algorithm is a seamless integration of novel change handling techniques into a diversity-centred sorting that has been deemed effective for static multi-objective optimisation \cite{jiang2017strength,jiang2016convergence}. SPEA/R \cite{jiang2017strength} relies on the use of a set of weight vectors which decompose the objective space of a problem into multiple subspaces and maintains population diversity by a diversity sorting mechanism across the subspaces, leading to different layers of population subsets: $L_1, L_2, \dots, L_s$ (in diversity and convergence decreasing order). SPEA/R starts population preservation for next round of evolution from layer $L_1$ until a population of desired size is achieved. This diversity-centred mechanism makes SPEA/R ideal (better than MOEA/D, see supplementary results) to be the search engine of DMO since it could effectively prevent loss of population diversity, a commonly observed issue in dynamic environments \cite{Goh2009dCOEA,Jiang2019SDP}. The framework of VARE is presented in \textbf{Algorithm \ref{alg:VARE}}. First, a set (with a size equal to population size) of uniformly-distributed reference directions is created (line 1) that is required to facilitate diversity-centred sorting and population prediction (see Fig. \ref{fig:vare}). Meanwhile, a initial parent population $P$ and a zero-vector $\pi$ are created (line 2--3), where $\pi$ is used to determine a mechanism in response to environmental changes (which will be detailed later). In the main while loop of the optimisation course, change detection is carried out at the beginning of each generation. If a change is detected, the evolved population for the last environment is then saved into an archive (line 6) and a change response is chosen to handle the new environment (line 7-14) and yield an offspring population $Q$, otherwise $Q$ is created via a genetic operation such as estimation of distribution \cite{Zhang2008rm}(line 16). After that, the diversity-centred sorting is applied to the union of $P$ and $Q$. At the end of every generation, the VARE probability $\pi$ of choosing certain change response mechanism is updated (line 19). In the following subsections, we will present the proposed change response mechanisms in great detail.

%%%% Algorithm 1 %%%%
\begin{algorithm}[t]
\caption{The VARE Framework}
\label{alg:VARE}
\KwIn {	stopping criteria,
	population size ({$N$});}

\KwOut {a set of PS approximations $\{\!P_1, P_2, \dots\}$;}

	Generate a uniform spread of $N$ reference directions: 
	$\{\lambda_1$, $\lambda_2$, $\dots$, $\lambda_N\}$ \;
	Initialise a random population $P=\{x^1,\dots,x^N\}$ \;
	Initialise VARE probability $\pi$ as a zero vector\;
	
	\While{Stopping criteria not met}{
		\eIf{Change detected }{
			Save $P$ to archive $A$\;
			\For{$i\gets 1$  \KwTo $N$}
			{
				\eIf{$\pi_i>rnd(0,1)$}{
					Generate an individual $q_i$ by VAR prediction (\textbf{Algorithm 2})\;
				}
				{
					Generate $q_i$ by environment-aware hypermutation (\textbf{Algorithm 3})\;
				}
			}
		$Q=\{q_1, q_2, \dots, q_N\}$\;
		}
		{
			Create $Q$ by genetic operation on $P$\;
		}
		Apply diversity-centred sorting \cite{jiang2017strength} on $P \cup Q$ to update $P$\;
		Update VARE probability $\pi$ by Eq.(\ref{eq:vare_prob})\;
}
\end{algorithm}
\setlength{\textfloatsep}{0pt}

\vspace{-5mm}
\subsection{Vector Autoregressive Population Prediction}
\vspace{-2mm}
VARE aims to build for each reference director $\lambda_i$ an autoregresor based on solutions found in past environments to predict a promising solution for the new environment. Each solution to a DMOP is a $n$-dimensional vector ($n$ decision variables), whose components are often correlated and non-separable from each other \cite{Huband2006WFG}. To account for the bidirectional influences between variables, VARE employs vector autoregression (VAR), which has been widely used in financial time-series forecasting applications \cite{geraci2018measuring}. Mathematically, the VAR model of lag order $l$, denoted as VAR($l$), for $n$ variables is given by:
%%% Figure 1 %%%%%
\begin{figure}[t]
\centering
\includegraphics[width=0.8\linewidth]{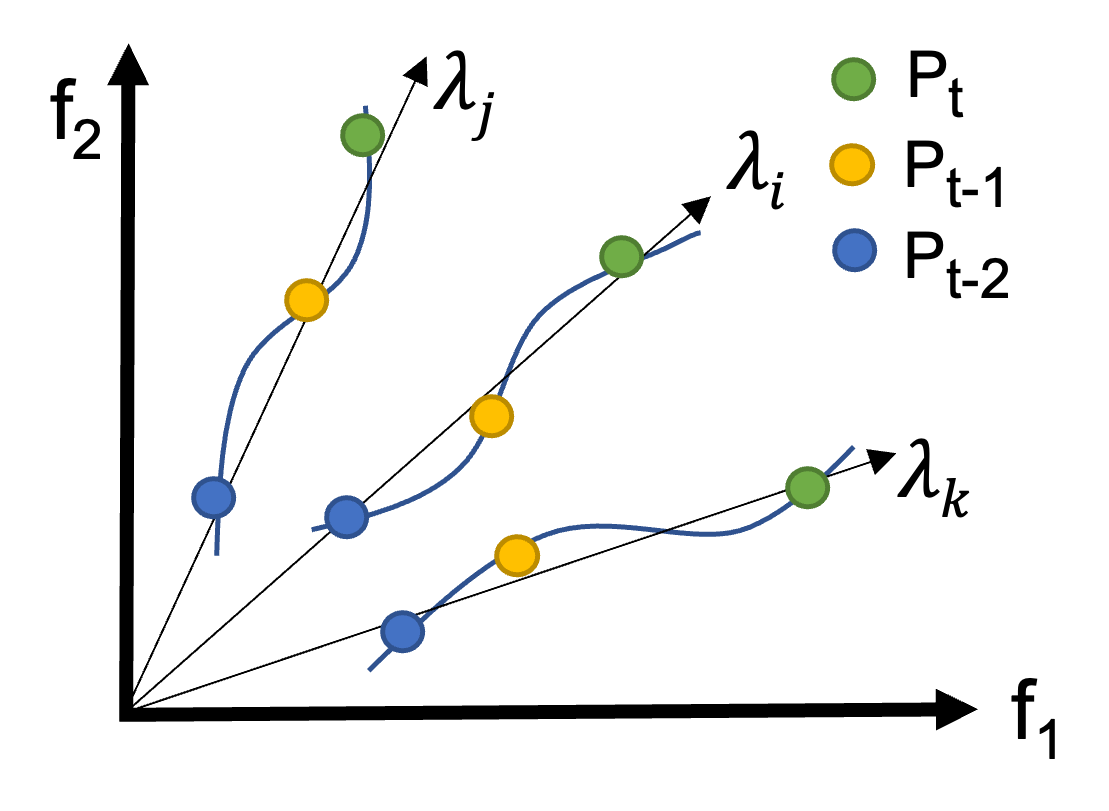} \\[-4mm]
\caption{Reference direction guided population prediction.}
\label{fig:vare}
\vspace{-0mm}
\end{figure}
\begin{align}
    x_{1,t}=&\alpha_{1}+\beta_{11}^1x_{1,t-1}+\cdots+\beta_{1l}^lx_{1,t-l}+\cdots\nonumber\\[-3pt]
    &+\beta_{nl}^1x_{n,t-1}+\cdots+\beta_{nl}^lx_{n, t-l}+\mu_{1}\nonumber\\[-3pt]
    x_{2,t}=&\alpha_{2}+\beta_{21}^1x_{2,t-1}+\cdots+\beta_{2l}^lx_{2,t-l}+\cdots\nonumber\\[-3pt]
    &+\beta_{nl}^1x_{n,t-1}+\cdots+\beta_{nl}^lx_{n, t-l}+\mu_{2}\nonumber\\[-5pt]
    \cdots\nonumber
    %x_{n,t}=\beta_{n0}+\beta_{n1}x_{n,t-1}+\cdots+\beta_{nl}x_{n,t-l}+\cdots+\gamma_{nl}x_{n-1,t-1}+\cdots+\gamma_{nl}x_{n-1, t-l}+\mu_{nt}\\
\end{align}
and written in vector form:
\begin{equation}
    x_t=\alpha +\beta^1 x_{t-1} +\beta^2 x_{t-2}+\cdots+\beta^l x_{t-l}+\mu
\end{equation}
where $\alpha=(\alpha_1,\dots, \alpha_n)^T$ and $\beta^k=(\beta_{ij}^k)_{1\le i, j\le n}$, $k=1, \dots, l$ are coefficients (vector/matrix) of the model and $\mu=(\mu_{1},\dots, \mu_{n})^T$ is an unobservable zero mean white noise vector process. Inference on coefficients involves estimating ($n+n^2l$) coefficients and this can be done by ordinary least squares. However, this dense parameterisation often leads to inaccuracies with regard to out-of-sample forecasting and structural inference, especially for higher-dimensional models \cite{kuschnig2021bvar}, which is the case with our study on population prediction for DMO. Thus, we propose to apply dimensionality reduction techniques to enable the use of VAR in a much lower dimensional space such that model parameterisation is efficient for population prediction.  

\begin{figure}
\centering
\setlength{\tabcolsep}{0pt}
\begin{tabular}{cc}
\includegraphics[width=4.4cm,height=4cm]{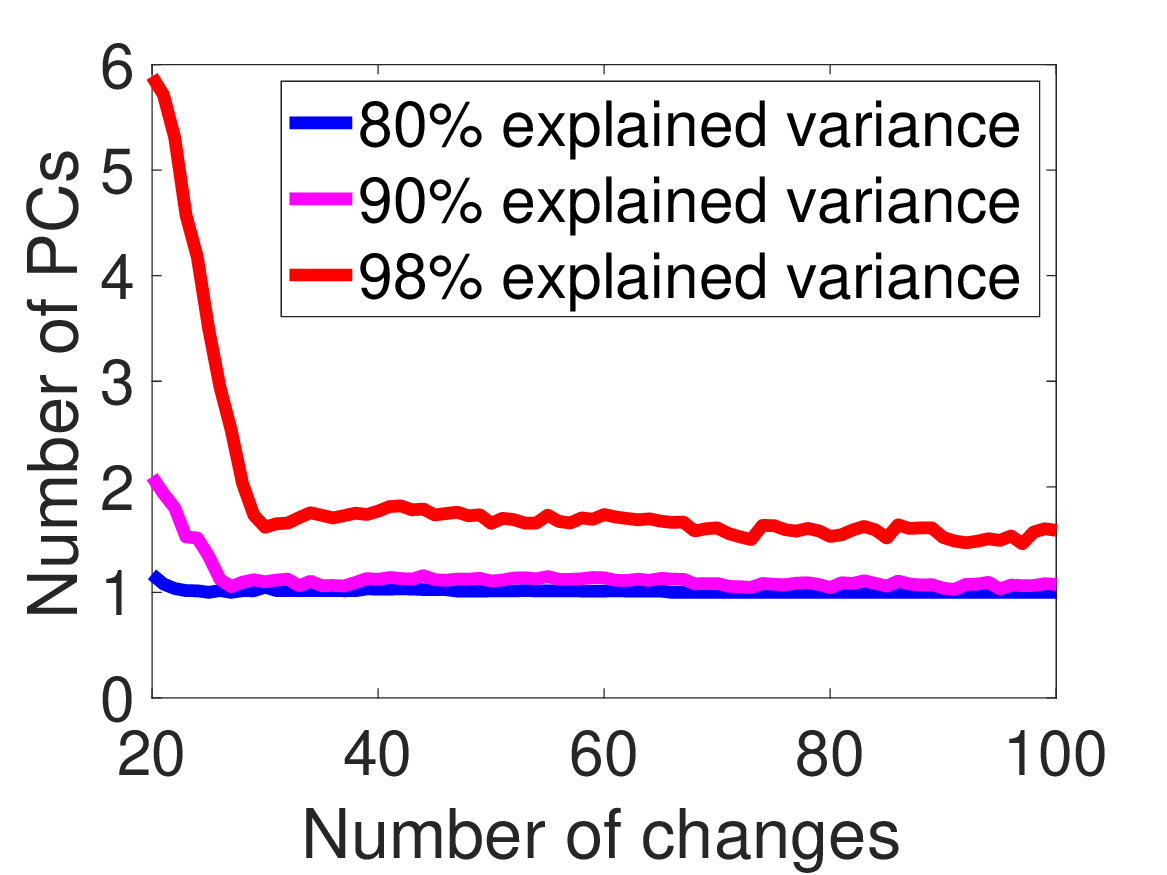} &
\includegraphics[width=4.4cm,height=4cm]{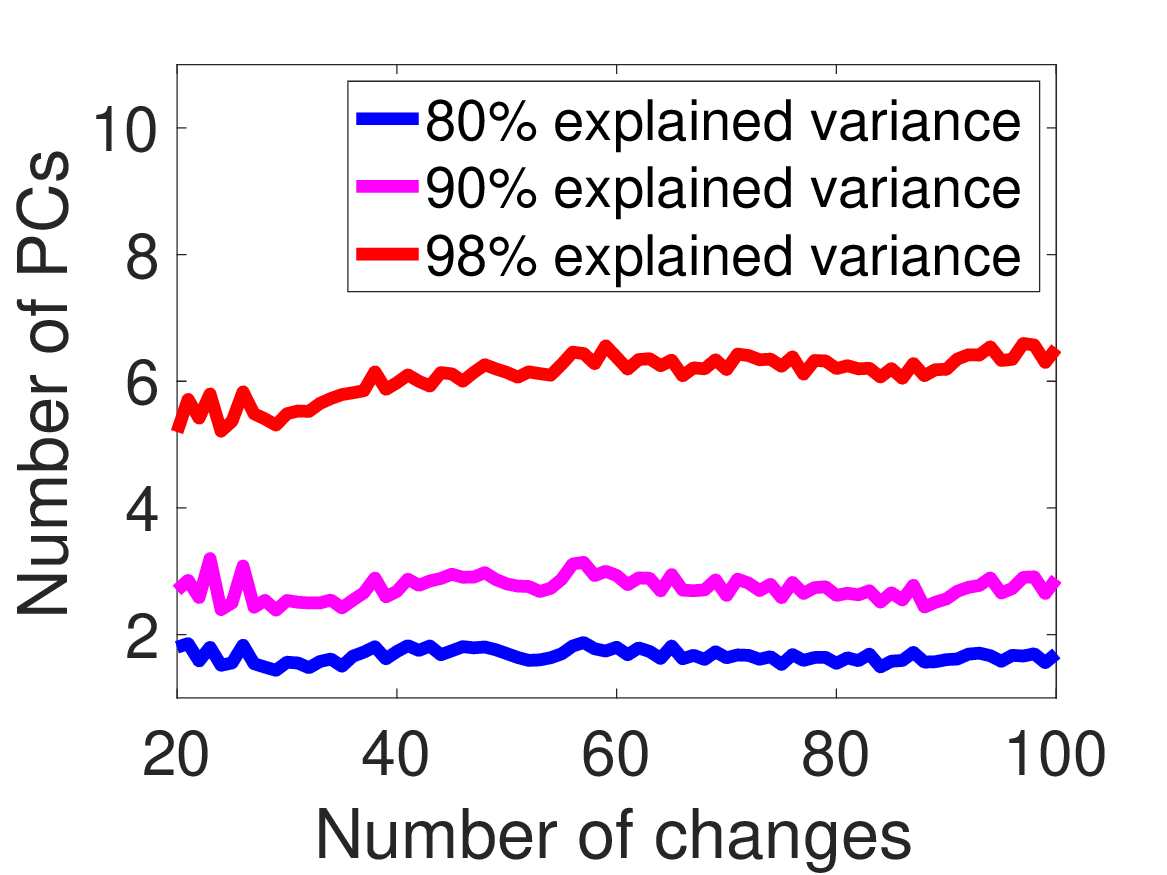} \\
(a) DF4 (2-objective)   & (b) DF10 (3-objective)\\
\end{tabular}
\caption{The average number of PCs needed for different levels of explained variance. Data is collected between 20th and 100th environmental change.}
\label{fig:num_pca}
\vspace{-2mm}
\end{figure}

\begin{figure}
\centering
\includegraphics[width=0.8\linewidth]{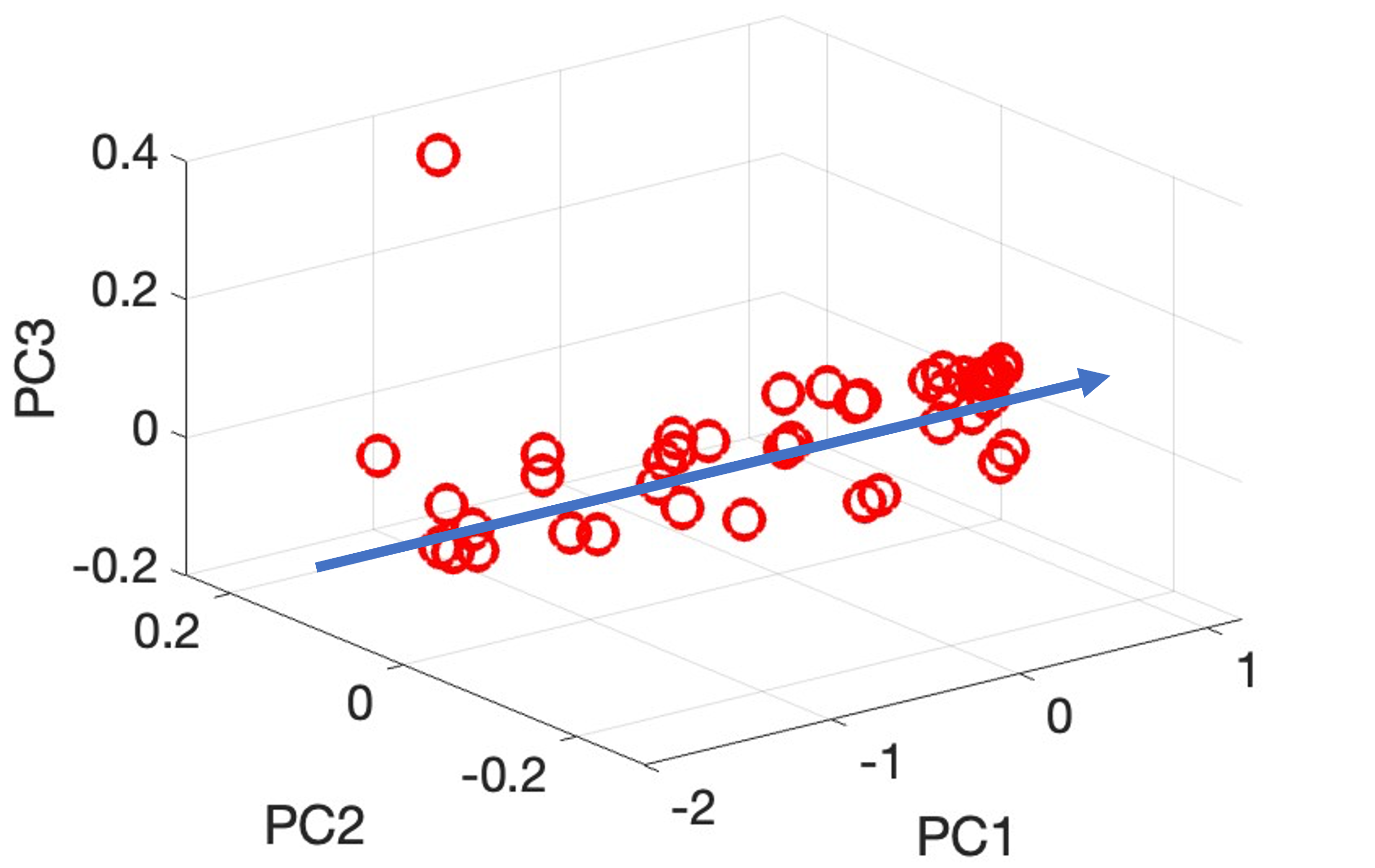}\\[-2mm]
\caption{Low-dimensional representation of 10-dimensional solutions obtained from the first 40 environmental changes for DF4 along the search direction $\lambda=(0, 1)$. The data can be maximally explained along the blue arrow, i.e., PC1.
% pc1(84%); pc2(12%); pc3(3%)
}
\label{fig:DF4_pca}
%\vspace{-2mm}
\end{figure}

Let $A_i=\{a_i^1, a_i^2,\dots, a_i^t\}$ (remember each $a_i$ is a $n\times 1$ vector) be all archived solutions associated with $\lambda_i$ up to the current environment. We want to map  any data point $x \in A_i$ from the $n $-D space to a much lower $k$-D ($k\! \ll\! n$) space, namely, 
\begin{equation}
%	x\!=\!(\alpha_1,\! \alpha_2, \!\dots, \!\alpha_n)^T\Longrightarrow  y\!=\!(\beta_1, \!\beta_2,\! \dots, \!\beta_k)^T 
 x\!=\!(x_1,\! x_2, \!\dots, \!x_n)^T\Longrightarrow  y\!=\!(y_1, \!y_2,\! \dots, \!y_k)^T 
\end{equation}
To this end, we choose the widely-used PCA for dimensionality reduction, although we recognise that other approaches such as t-SNE and UMAP may also suffice with higher computational costs but they are unsuitable for the reconstruction of $x$ from $y$, which is required in this work. PCA requires constructing  from $A_i$ a  $n \times n$ scatter matrix $C=\sum_{j=1}^{t}(a_i^j-\bar{A}_i)(a_i^j-\bar{A}_i)^T$, where  $\bar{A}_i$ is the mean of $A_i$, followed by computing its top $k$ eigenvectors (principal components, i.e. PC) $V_k= (v_1, v_2, \dots, v_k)$ such that $C\approx V_k\Sigma_kV_k^T$, where $\Sigma_k$ is the diagonal matrix of the $k$ eigenvalues $(\sigma_1, \sigma_2,\dots, \sigma_k)$ corresponding to $V_k$. $k$ is determined by selecting the smallest values of $k$ that can explain 80 \% of the variance of the data $A_i$. We have found that this setting often leads to $k=1$ or 2, a significant reduction of dimensionality from a $n$-D ($n\ge 10$) space (see Fig. \ref{fig:num_pca}). Fig. \ref{fig:DF4_pca} exemplifies that solutions found in previous environments can be effectively represented in a much lower dimensional space through PCA.

After efficient PCA on small samples in our study, VAR is built in the reduced space of time-series solutions collected along reference direction $\lambda_i$. We use a Bayesian estimation method \cite{kuschnig2021bvar}, which is generally faster than ordinary least squares, to further speed up parameter inference for VAR in the reduced space. Once such a model is built, it is used to make one-step prediction $\tilde{a_i}^{t+1}$ in the reduced space and then $\tilde{a_i}^{t+1}$ is mapped to the original decision space using
\begin{equation}
	q_i = \tilde{a_i}^{t+1} V_k^T+\bar{A}_i,
	\label{eq:var}
\end{equation}
where $q_i$, reconstructed from $\tilde{a_i}^{t+1}$, is the predicted solution for the new environment $t+1$. The procedure of VAR based prediction is described in \textbf{Algorithm \ref{alg:VAR}}.
%%% Figure 1 %%%%%

%%% Figure 1 %%%%%

\vspace{-2mm}
\subsection{Environment-Aware Hypermutation}
The prediction approach described above requires sufficient data to build a VAR model. Other change response mechanisms are needed when data is small in early environmental changes% until a sufficient number of past solutions are accumulated for VAR modelling
. In addition, VAR may not always be effective, especially when there are environmental changes for which prediction-based approaches are unsuitable \cite{Jiang2019SDP}. For these reasons, we have developed environment-aware hypermuation (EAH) in conjunction with VAR prediction to handle environmental changes.

%%%% Algorithm 2 %%%%
\begin{algorithm}[t]
	\caption{VAR Based Prediction}
	\label{alg:VAR}
	\KwIn {
		archive ($A$),
		reference direction ({$\lambda_i$}),
		lag length({$l$});}
	
	\KwOut {a predicted individual $q_i$ for $\lambda_i$;}
	
	Identify archived individuals $A_i=\{a_i^1, a_i^2,\dots, a_i^t\}$ from $A$ that are associated with $\lambda_i$\;
	Apply $PCA(A_i)$ to get $k$ principal components $V_k$ and the mapping $\tilde{\mathcal{A}_i}$ of $A_i$ in this $k$-D space\;
	Learn a VAR($l$) model from $\tilde{\mathcal{A}_i}$\;
	Make one-step prediction $\tilde{a_i}^{t+1}$ using VAR($l$)\;
	Reconstruct an image $q_i$ of $\tilde{a_i}^{t+1}$ in the original variable space using Eq.(\ref{eq:var})\;
\end{algorithm}
%%% Figure 2 %%%%%
\begin{figure}[t]
\centering
\setlength{\tabcolsep}{0pt}
\begin{tabular}{cc}
\includegraphics[width=4.4cm,height=4cm, trim=12 3 11 3,clip]{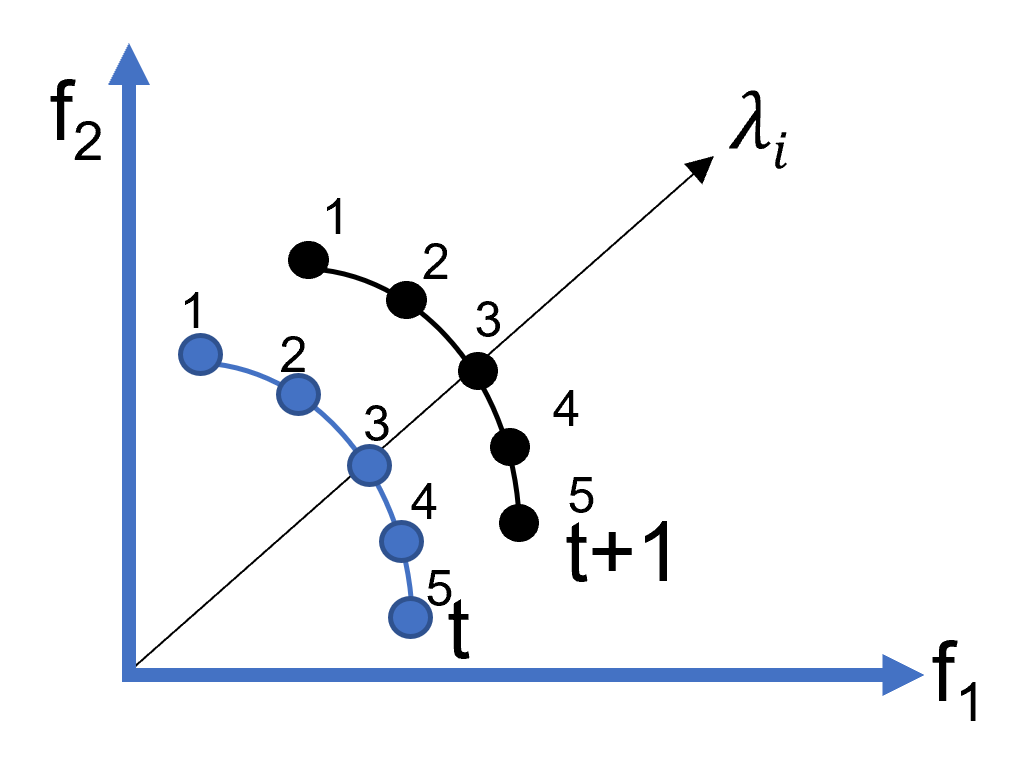} &
\includegraphics[width=4.4cm,height=4cm, trim=12 3 12 3,clip]{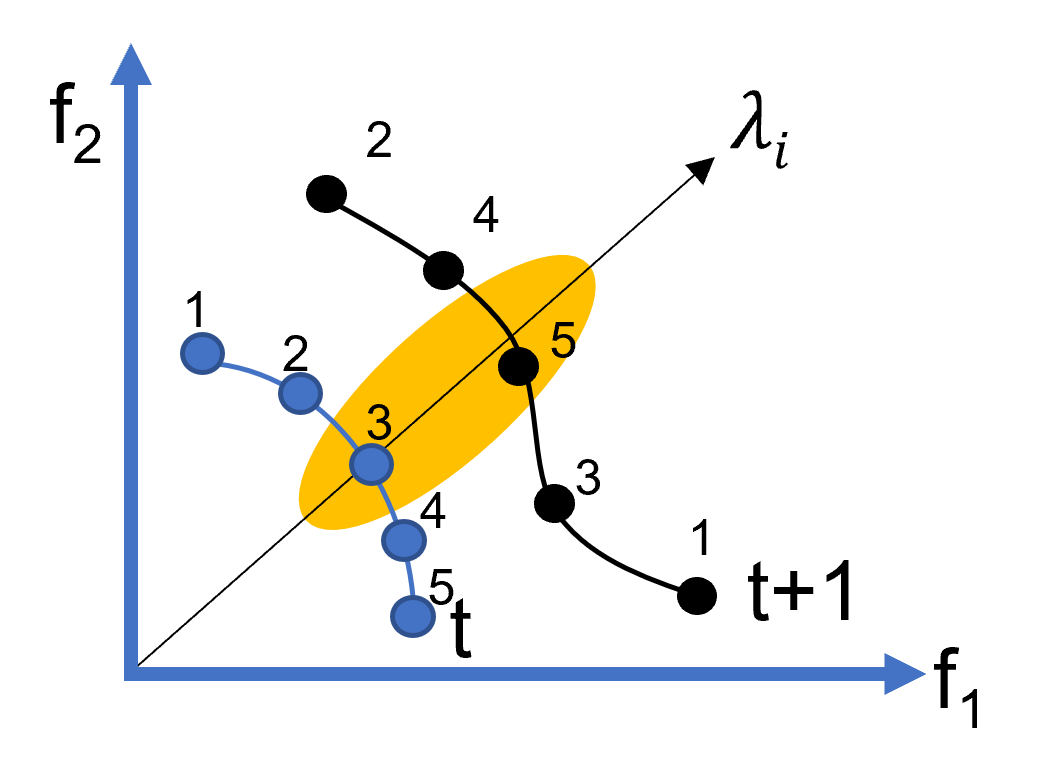} \\[-2mm]
(a) No change in PS   & (b) Change in PS\\
\end{tabular}
\caption{Illustration of two scenarios of changes, suggesting estimation of severity of change should account for change information in both objective and decision spaces.}
\label{fig:hymute}
\end{figure}

Hypermutation is a strategy widely used in dynamic environments to augment population diversity, by increasing the amount of mutation to existing solutions,  when a change is detected. Existing hypermutation strategies are often (partially) blind to environments and thus cannot effectively determine the best amount of mutation for diversity increase, since too much mutation can have the same effect as random solutions whereas too little mutation makes no significant difference to population diversity. In the EDMO literature, there are two types of hypermutation: ({\romannumeral 1}) environment-blind hypermutation such as D-NSGA-II-B \cite{Deb2007DNSGA2} that simply changes the distribution index $\eta$ of polynomial mutation to a predefined value; and ({\romannumeral 2}) partially blind hypermutation such as \cite{Li2022,Sahmoud2019severity} that relies on objective variation to determine the amount of mutation in population. While ({\romannumeral 2}) is more rational than ({\romannumeral 1}), it can produce negative results in some cases. For example, Fig.~\ref{fig:hymute}(a) shows a scenario where the PS remains the same but objective values are shifted by a constant after an environmental change. In other words, solutions found at environment $t$ are still perfectly resuable at environment $t+1$ although there is objective variation. In this case, hypermutation ({\romannumeral 2}) would have adverse effects on population diversity %and lead to decreased quality of hypermutated solutions that are included in the population 
for the new environment. 

Here, we have developed EAH that is well aware of environmental changes in both decision and objective space. EAH collects information from both spaces to estimate the severity of change, which is then used to determine the value of $\eta$ in polynomial mutation. In objective space, EAH calculates the average of relative objective difference across all $M$ objectives and the population (size $N$), denoted $\Delta_F$ as follows:
\begin{equation}
\Delta_F = {\frac{1}{MN}} \sum_{i=1}^{N}\sum_{j=1}^{M}\frac{\left|f_j(x_i,t\!+\!1)\!-\!f_j(x_i,t)\right|}{\left|f_j(x_i,t)\right|+\epsilon},
\end{equation}
where $\epsilon=10^{-6}$ is used to keep division valid.
In decision space, for each solution $x_i\in P_t$ associated with reference direction $\lambda_i$ at environment $t$, EAH identifies for $x_i$ the best partner $\hat{x}_i\in \hat{P}_t$ (reevaluated $P_t$ at environment $t+1$) associated with the same reference direction $\lambda_i$ (see Fig.~\ref{fig:hymute}(b)). Then, we test whether $\hat{x}_{i,j}$ significantly differs from $x_{i,j}$ by conducting a right-tailed T-test (confidence level 0.95) on the mean $\mu$ of $\left|\hat{x}_{i,j}-x_{i,j}\right|$ as follows:
\begin{equation}
H_0: \mu= 0 \mbox{ vs } H_A: \mu > 0
\end{equation}
%the mean of $\delta_{i,j}$ significantly differs from  $\delta_{i,j}=\left|\hat{x}_{i,j}-x_{i,j}\right|$, for $1\le i \le N$ and $1 \le j \le n$, and we  conduct right-tailed T-test
In addition, the average of relative decision difference, denoted as $\Delta_X$, is calculated across all $n$ decision variables and the population by:  
\begin{equation}
    \Delta_X={\frac{1}{Nn}} \sum_{i=1}^{N}\sum_{j=1}^{n}\frac{\left|\hat{x}_{i,j}-x_{i,j}\right|}{\left|x_{i,j}\right|+\epsilon}.
\end{equation}
After that, $\Delta_F$ and $\Delta_X$ are integrated to calculate the distribution index $\eta$:
\begin{equation}
\setlength{\jot}{-5pt}
\eta\!\!=\!\!\left \{ \begin{array}{ll}
\!\!20, & \mbox{if $H_0$ is accepted;} \\[1mm]
\!\!20\!*\!max(e^{-(\Delta_F+\Delta_X)},0.1), & \mbox{otherwise}.
\end{array}
\right.
    \label{eq:eta}
\end{equation}
This equation restricts $\eta$ to the range [2, 20], where $\eta=20$ is a common choice in static multi-objective optimisation and $\eta \ge 2$ is inspired by the study \cite{Deb2007DNSGA2} to make sure that hypermutated solutions are not too random. This way, the situation presented in Fig.~4(a) can be effectively addressed by enabling small mutation (via large $\eta$) due to the acceptance of $H_0$ in the case of no PS change. The pseudocode of EAH is presented in \textbf{Algorithm \ref{alg:EAH}}.

% \begin{gather}
% 	\Delta_F = {\frac{1}{MN}} \sum_{i=1}^{N}\sum_{j=1}^{M}\left|\frac{f_j(x_i,t\!+\!1)\!-\!f_j(x_i,t)}{f_j(x_i,t)+\epsilon}\right|\\
% 	\Delta_X={\frac{1}{Nn}} \sum_{i=1}^{N}\sum_{j=1}^{n}\left|\frac{\hat{x}_{i,j}-x_{i,j}}{x_{i,j}+\epsilon}\right|\\
% 	\eta=20*max(e^{-(\Delta_F+\Delta_X)},0.1)
% 	\label{eq:eah}
% \end{gather}

%%%% Algorithm 3 %%%%
\begin{algorithm}[t]
	\caption{Environment Aware Hypermutation}
	\label{alg:EAH}
	\KwIn {	
		archive ($A$),
		reference direction ({$\lambda_i$});}
	
	\KwOut {a hypermuted individual $q_i$ for $\lambda_i$;}
	
		Identify archived individuals for the last two environments $\{a_i^{t-1}, a_i^t\}$ from $A$ that are associated with $\lambda_i$\;
		Compute mutation index $\eta$ according to Eq.(\ref{eq:eta})\;
		
		Hypermutate $a_i^t$ with $\eta$ to create  individual $q_i$\;
\end{algorithm}

\vspace{-1mm}
\subsection{Adaptive Change Response}
As mentioned earlier, EAH can be used in early environmental changes when there is not enough data for VAR modelling and when VAR prediction %VAR models cannot be built due to lack to data but also replace VAR prediction when the latter is 
is ineffective to handle certain environmental changes that exhibit little predictable patterns. Thus, we have developed an adaptive change response scheme that selects either VAR prediction or EAH in an evolutionary manner. To do so, we define $\pi_i$, the probability of selecting VAR prediction as the change response mechanism for reference direction $\lambda_i$, as follows:
\begin{equation}
	\pi_i =\frac{\rho_{i, p}}{\rho_{i, p}+\rho_{i,m}},
	\label{eq:vare_prob}
\end{equation}
where  $\rho_{i, s}$ is the success rate of generating by the strategy $s$ (which denotes either VAR prediction or EAH) an individual that is preserved in each initial population of the last $L$ environmental changes. Eq.~\ref{eq:vare_prob} means that, in the event of an environmental change, the solution associated with reference direction $\lambda_i$ is generated by VAR prediction with a probability of $\pi_i$ and by EAH with a probability of $1-\pi_i$. This adaptive scheme has a tunable parameter $L$. However, this parameter is unsurprisingly coupled with the lag order $l$ of the VAR model ($L=l$), since the optimal $l$ for VAR prediction means the last $l$ environments affect most the predicted solutions for the new environment. Our sensitivity analysis detailed later confirms the above hypothesis, i.e., $L=l$ is the best setting for VARE. 

\subsection{Complexity Analysis of VARE}
Here we analyse the computational complexity of VARE for one environmental change. The main time complexity comes from the proposed change response mechanisms. VAR prediction in this work requires PCA for dimentionality reduction, which has a complexity of $\mathcal{O}(Kn*min(K, n)+n^3)$ \cite{elgamal2015analysis} where $K$ is the number of environmental changes occurred so far, and $n$ is the number of decision variables. VAR modelling based on Bayesian estimation techniques has a complexity of $\mathcal{O}(k^3)$ \cite{korobilis2020sign}, where $k$ is the reduced dimension after PCA ($k$ is usually small in this work). Both PCA and VAR prediction are used for all $N$ reference directions (population size). Therefore their total computational cost is mainly from PCA, i.e., $\mathcal{O}(N(K^2n+n^3))$ (since $K$ is usually larger than $n$). EAH requires the association of solutions to reference directions, which has a complexity of $\mathcal{O}(N^2)$. Diversity-centred sorting  at the end of each generation has a complexity of $\mathcal{O}(N^2)$ \cite{jiang2017strength}. The complexity of other procedures is negligible. Note that dimension reduction happens only once in each environmental change and the fact that PCA is quite fast for small DMO data, thus the average generational time complexity of VARE comes from diversity-centred sorting, i.e., $\mathcal{O}(N^2)$.

\section{Experimental Settings} 
\label{sect:emp}

\subsection{Problems and Algorithms} 
Test problems and popular algorithms are used to evaluate the proposed algorithm, which are described in the following paragraphs in detail.
\subsubsection{Test Problems}
A range of synthetic DMOPs from DF \cite{Jiang2018DF} and FDA \cite{Farina2004deb} test suites are used in this paper. These two test suites have been widely adopted for facilitating algorithm analysis and development. The DF test suite, which was proposed for IEEE CEC2018 Competition on DMO, includes a number of well-defined problems with a diverse set of dynamic features. In addition, three triobjective  problems, i.e., F8 \cite{Zhou2014PPS} and FDA4--FDA5 \cite{Farina2004deb}, are considered in this study.
All the test problems have no less than 10 decision variables. A range of values for severity of change $n_t \in \{5, 10, 20\}$ and frequency of change $\tau_t\in \{10, 20, 30\}$ are investigated. 

\subsubsection{Compared Algorithms}
Several popular algorithms developed recently are considered for comparison to demonstrate the effectiveness and efficiency of the proposed algorithm. They are listed as follows:
\begin{itemize}
    \item PPS \cite{Zhou2014PPS}: This algorithm proposed a population prediction strategy that uses an autoregressive model to predict the movement of population centroids in dynamic environments and estimates the manifold of PS/PF represented by the population. RM-MEDA \cite{Zhang2008rm} is used for reproduction. We use for this algorithm the same parameter setting as its original paper \cite{Zhou2014PPS}.

    \item SGEA \cite{Jiang2016SGEA}: This algorithm proposed to address dynamic changes in an steady-state and generational manner. It combines the advantages of steady-state evolution in fast population update and generational evolution in diversity maintenance, thus it is able to quickly detect environmental changes and make rapid responses. We use the original parameter setting for this algorithm \cite{Jiang2016SGEA}. 

    \item Tr-RM-MEDA \cite{Jiang2017transfer}: This is machine learning based algorithm for DMO. Tr-RM-MEDA applies transfer learning techniques to map archived solutions into a latent space from which an initial population for a new environment is generated, a computationally intensive process that requires solution sampling. It uses RM-MEDA \cite{Zhang2008rm} as the reproduction operator. For fair comparison, we limited the sample size to be same as population size so that the total number of function evaluations are equal to that of the other algorithms in each time window of environmental change.

    \item MOEA/D-SVR \cite{Cao2019MOEADSVR}: This algorithm builds on the popular decomposition-based algorithm MOEA/D to predict a solution for each weight vector of MOEA/D after an environmental change. The prediction relies on support vector regressors (SVR) with nonlinear kernels. For each weight vector, it requires as many SVRs as the number of decision variables, making this algorithm computationally intensive. The parameter setting remains the same as its original paper \cite{Rong2020multimodel}.

    \item VARE \footnote{A MATLAB implementation of VARE is available upon request.}: This is the proposed algorithm in this paper. The lag order $l=5$ is used. Like PPS and Tr-RM-MEDA, VARE employs RM-MEDA \cite{Zhang2008rm} as the variation operator for reproduction. 
\end{itemize}
In the experiment, all the algorithms used a population size of $N=100$ for biobjective problems and $N=105$ for triobjective problems in order to be consistent with %weight or reference generation in 
MOEA/D-SVR and VARE. 

\subsection{Performance Indicators}
In our empirical studies, we adopt the following performance 
indicators.

\subsubsection{Mean Inverted Generational Distance (MIGD)} 
%MIGD \cite{Jiang2016JY} is adapted from the static performance indicator IGD \cite{Zhang2008rm} that measures both the convergence and diversity of solutions found by an algorithm. 
Let $P_t$ be a set of $n_{P_t}$ uniformly distributed points in the true PF, and $P_t^*$ be an approximation of the PF, at time $t$. The MIGD \cite{Jiang2016JY} is calculated as follows:
\begin{equation}
MIGD=\frac{1}{T}\sum\limits_{i=1}^{T}IGD(P_t^*,P_t)=\frac{1}{T} \sum\limits_{i=1}^{T}\sum\limits_{i=1}^{n_{P_t}}{\frac{d_t^i}{|P_t|}},\\
\end{equation}  
where $d_t^i$ is the Euclidean distance between the $i$-th member in $P_t$ and its nearest member in $P_t^*$. A set of around 10,000 points uniformly sampled from the true PF is used for MIGD calculation.

\subsubsection{Mean Hypervolume (MHV)}
The MHV \cite{Jiang2016SGEA} is a modification of the static hypervolume (HV) measure that computes the volume of the area dominated by the obtained $P_t^*$:
\begin{equation}
{MHV}=\frac{1}{T}\sum\nolimits_{i=1}^{T}HV_t(P_t^*),
\end{equation}
where $HV(S)$ is the hypervolume of a set $S$. The reference point for the computation of hypervolume is $(z_1+0.1, \cdots, z_M+0.1)$, where $z_j$ is the maximum value of the $j$-th objective of the true PF at time $t$ and $M$ is the number of objectives.

Note that, we applied max-min normalisation to $P_t$ and $P_t^*$ before the calculation of IGD and HV, where the maximum and minimum values refer to the lower and upper bounds of problem objectives at time $t$, respectively. This ensures that IGD values obtained in different environments are comparable and equally treated in MIGD regardless of the scale of objectives (the range of objectives may vary significantly in dynamic environments).

\subsubsection{Runtime}
Execution time is another measure adopted in this paper to evaluate the efficiency of different algorithms. To do so, we implemented all the compared algorithms in MATLAB 2022a based on these algorithms' original implementation. All of them were run independently 30 times on a Linux computer with Intel Xeon Gold 6138 2.00Ghz processors. Average runtime and standard deviations are reported.

% Table 1 %
\begin{table*}[htbp]
    % \addtolength{\tabcolsep}{-4pt}
    % \setlength\extrarowheight{-3pt}
    % \renewcommand{\arraystretch}{-0.2}
  \centering
  \caption{Mean (standard deviation) values of MIGD and MHV (with background colour) obtained by five algorithms, with best ones in boldface}
    \begin{tabular}{lllllll}
    \toprule
    No.   & Prob.  & VARE  & SGEA  & PPS   & Tr-RM-MEDA & MOEA/D-SVR \\
    \midrule
    \multirow{2}[2]{*}{1} & \multirow{2}[2]{*}{DF1} & 1.64E-2(2.22E-3) & 7.18E-2(7.01E-3)$\ddag$ & 3.23E-1(3.36E-2)$\ddag$ & \textbf{8.91E-3(3.64E-4)} & 7.43E-2(6.16E-3)$\ddag$ \\
          &       & \cc6.30E-1(3.43E-3) & \cc5.47E-1(9.04E-3)$\ddag$ & \cc3.31E-1(1.65E-2)$\ddag$ & \cc\textbf{6.45E-1(5.42E-4)} & \cc5.59E-1(5.85E-3)$\ddag$ \\
    \midrule
    \multirow{2}[2]{*}{2} & \multirow{2}[2]{*}{DF2} & 6.46E-2(4.36E-3) & 5.82E-2(4.62E-3) & 2.15E-1(1.85E-2)$\ddag$ & \textbf{4.17E-3(4.10E-5)} & 4.18E-2(3.79E-3) \\
          &       & \cc7.56E-1(7.97E-3) & \cc7.78E-1(5.08E-3) & \cc5.57E-1(1.84E-2)$\ddag$ & \cc\textbf{8.70E-1(1.01E-4)} & \cc8.25E-1(2.95E-3) \\
    \midrule
    \multirow{2}[2]{*}{3} & \multirow{2}[2]{*}{DF3} & \textbf{8.92E-3(8.05E-4)} & 4.72E-2(6.22E-2)$\ddag$ & 1.06E-1(4.70E-2)$\ddag$ & 1.18E-1(3.24E-3)$\ddag$ & 4.53E-1(8.51E-3)$\ddag$ \\
          &       & \cc\textbf{6.06E-1(1.43E-3)} & \cc5.66E-1(4.42E-2)$\ddag$ & \cc4.89E-1(3.45E-2)$\ddag$ & \cc4.59E-1(5.06E-3)$\ddag$ & \cc2.37E-1(5.17E-3)$\ddag$ \\
    \midrule
    \multirow{2}[2]{*}{4} & \multirow{2}[2]{*}{DF4} & \textbf{2.87E-2(3.74E-4)} & 3.30E-2(8.20E-4)$\ddag$ & 3.26E-2(1.15E-3)$\ddag$ & 2.64E-1(9.15E-3)$\ddag$ & 5.20E-2(7.39E-4)$\ddag$ \\
          &       & \cc\textbf{8.65E-1(9.33E-4)} & \cc8.57E-1(1.18E-3)$\ddag$ & \cc8.49E-1(1.95E-3)$\ddag$ & \cc5.21E-1(9.62E-3)$\ddag$ & \cc8.12E-1(9.75E-4)$\ddag$ \\
    \midrule
    \multirow{2}[2]{*}{5} & \multirow{2}[2]{*}{DF5} & \textbf{1.17E-2(1.47E-3)} & 2.04E-1(5.36E-2)$\ddag$ & 3.51E-1(8.24E-2)$\ddag$ & 6.48E-2(5.26E-3)$\ddag$ & 6.23E-2(4.89E-3)$\ddag$ \\
          &       & \cc\textbf{6.87E-1(2.29E-3)} & \cc4.91E-1(2.47E-2)$\ddag$ & \cc3.34E-1(3.46E-2)$\ddag$ & \cc6.05E-1(6.67E-3)$\ddag$ & \cc6.16E-1(4.71E-3)$\ddag$ \\
    \midrule
    \multirow{2}[2]{*}{6} & \multirow{2}[2]{*}{DF6} & \textbf{2.79E+0(9.48E-1)} & 2.91E+0(3.92E-1)$\ddag$ & 7.73E+0(6.50E-1)$\ddag$ & 6.41E+0(2.54E-1)$\ddag$ & 5.47E+0(2.57E-1)$\ddag$ \\
          &       & \cc\textbf{2.66E-1(3.35E-2)} & \cc1.69E-1(3.64E-2)$\ddag$ & \cc9.31E-3(5.18E-3)$\ddag$ & \cc1.21E-2(5.87E-3)$\ddag$ & \cc1.42E-1(4.20E-3)$\ddag$ \\
    \midrule
    \multirow{2}[2]{*}{7} & \multirow{2}[2]{*}{DF7} & 2.18E-2(2.81E-3) & 5.32E-2(5.56E-3)$\ddag$ & 4.01E-2(6.82E-3)$\ddag$ & \textbf{1.35E-2(1.42E-3)} & 2.43E-1(3.24E-3)$\ddag$ \\
          &       & \cc8.87E-1(2.32E-3) & \cc8.47E-1(7.63E-3)$\ddag$ & \cc8.51E-1(9.95E-3)$\ddag$ & \cc\textbf{9.05E-1(1.57E-3)} & \cc6.13E-1(4.51E-3)$\ddag$ \\
    \midrule
    \multirow{2}[2]{*}{8} & \multirow{2}[2]{*}{DF8} & 1.36E-2(3.92E-4) & 1.69E-2(2.68E-4)$\ddag$ & \textbf{1.09E-2(4.73E-4)} & 1.09E-1(4.50E-3)$\ddag$ & 2.30E-1(1.07E-2)$\ddag$ \\
          &       & \cc\textbf{7.43E-1(8.03E-4)} & \cc7.43E-1(4.09E-4)$\ddag$ & \cc7.39E-1(7.81E-4) & \cc6.24E-1(6.77E-3)$\ddag$ & \cc6.40E-1(8.85E-3)$\ddag$ \\
    \midrule
    \multirow{2}[2]{*}{9} & \multirow{2}[2]{*}{DF9} & \textbf{1.19E-1(1.78E-2)} & 1.91E-1(3.09E-2)$\ddag$ & 5.19E-1(6.54E-2)$\ddag$ & 1.22E-1(5.82E-2)$\dag$ & 4.68E-1(1.51E-2)$\ddag$ \\
          &       & \cc4.80E-1(1.89E-2) & \cc4.09E-1(2.37E-2)$\ddag$ & \cc2.24E-1(2.50E-2)$\ddag$ & \cc\textbf{4.84E-1(8.06E-3)}$\dag$ & \cc3.07E-1(1.08E-2)$\ddag$ \\
    \midrule
    \multirow{2}[2]{*}{10} & \multirow{2}[2]{*}{DF10} & 1.08E-1(3.48E-3) & \textbf{6.79E-2(1.93E-3)} & 1.68E-1(7.08E-3)$\ddag$ & 1.09E-1(1.87E-3)$\dag$ & 9.84E-1(1.40E-2)$\ddag$ \\
          &       & \cc8.71E-1(5.33E-3) & \cc\textbf{9.27E-1(4.54E-3)} & \cc7.38E-1(9.19E-3)$\ddag$ & \cc8.57E-1(4.37E-3)$\ddag$ & \cc1.97E-1(1.51E-2)$\ddag$ \\
    \midrule
    \multirow{2}[2]{*}{11} & \multirow{2}[2]{*}{DF11} & 1.14E+0(1.36E-3) & \textbf{1.13E+0(5.80E-4)} & 1.15E+0(1.03E-3)$\dag$ & 1.15E+0(7.12E-4)$\ddag$ & 1.24E+0(2.72E-3)$\ddag$ \\
          &       & \cc9.10E-2(1.12E-3) & \cc\textbf{1.01E-1(3.93E-4)} & \cc8.03E-2(1.18E-3)$\ddag$ & \cc8.93E-2(1.71E-3)$\ddag$ & \cc4.93E-2(8.68E-4)$\ddag$ \\
    \midrule
    \multirow{2}[2]{*}{12} & \multirow{2}[2]{*}{DF12} & \textbf{1.20E-1(3.08E-3)} & 2.41E-1(4.15E-3)$\ddag$ & 2.37E-1(2.88E-3)$\ddag$ & 2.19E-1(5.45E-3)$\ddag$ & 8.67E-1(5.07E-3)$\ddag$ \\
          &       & \cc\textbf{1.26E+0(2.35E-3)} & \cc1.18E+0(1.27E-2)$\ddag$ & \cc1.13E+0(1.09E-2)$\ddag$ & \cc1.22E+0(5.99E-3)$\ddag$ & \cc7.81E-1(3.06E-2)$\ddag$ \\
    \midrule
    \multirow{2}[2]{*}{13} & \multirow{2}[2]{*}{DF13} & 1.45E-1(4.18E-3) & \textbf{1.13E-1(5.81E-3)} & 2.06E-1(9.14E-3)$\ddag$ & 1.74E-1(5.04E-3)$\ddag$ & 3.35E-1(1.88E-2)$\ddag$ \\
          &       & \cc7.20E-1(8.31E-3) & \cc\textbf{8.04E-1(8.53E-3)} & \cc6.33E-1(1.52E-2)$\ddag$ & \cc7.18E-1(7.89E-3)$\dag$ & \cc6.63E-1(4.19E-2)$\ddag$ \\
    \midrule
    \multirow{2}[2]{*}{14} & \multirow{2}[2]{*}{DF14} & \textbf{8.23E-2(1.92E-3)} & 1.05E-1(7.11E-3)$\ddag$ & 1.91E-1(1.74E-2)$\ddag$ & 1.07E-1(3.34E-3)$\ddag$ & 2.75E-1(8.91E-3)$\ddag$ \\
          &       & \cc\textbf{9.17E-1(3.78E-3)} & \cc8.65E-1(1.28E-2)$\ddag$ & \cc7.03E-1(2.99E-2)$\ddag$ & \cc8.80E-1(7.87E-3)$\ddag$ & \cc6.21E-1(1.47E-2)$\ddag$ \\
    \midrule
    \multirow{2}[2]{*}{15} & \multirow{2}[2]{*}{F8} & \textbf{1.04E-1(6.18E-3)} & 1.07E-1(2.04E-2) & 3.59E-1(1.77E-2) & 6.23E-1(3.21E-2) & 6.85E-1(4.21E-2) \\
          &       & \cc\textbf{5.91E-1(1.63E-2)} & \cc5.78E-1(4.29E-2) & \cc1.69E-1(1.40E-2) & \cc1.63E-1(7.24E-3) & \cc2.25E-1(1.55E-2) \\
    \midrule
    \multirow{2}[2]{*}{16} & \multirow{2}[2]{*}{FDA4} & \textbf{1.71E+0(7.24E-3)} & 1.73E+0(3.89E-3)$\dag$ & 1.87E+0(1.52E-2)$\ddag$ & 2.05E+0(2.53E-2)$\ddag$ & 1.77E+0(6.90E-4)$\ddag$ \\
          &       & \cc4.00E-1(4.64E-3) & \cc3.66E-1(6.20E-3)$\ddag$ & \cc2.97E-1(9.44E-3)$\ddag$ & \cc\textbf{4.35E-1(2.19E-3)} & \cc2.83E-1(1.41E-3)$\ddag$ \\
    \midrule
    \multirow{2}[2]{*}{17} & \multirow{2}[2]{*}{FDA5} & 7.05E-2(5.97E-3) & 5.32E-2(2.59E-3) & 8.78E-2(7.76E-3)$\ddag$ & \textbf{4.09E-2(1.29E-3)} & 1.44E-1(7.69E-3)$\ddag$ \\
          &       & \cc\textbf{5.71E-1(1.72E-2)} & \cc4.90E-1(2.24E-2) & \cc3.51E-1(1.95E-2) & \cc5.48E-1(6.39E-3) & \cc3.38E-1(2.22E-2) \\
    \midrule
    \multicolumn{2}{l}{Overall rank*} & 1 (1.47) & 2 (2.53) & 4 (3.82) & 3 (2.59) & 5 (4.12)\\
    \bottomrule
    \multicolumn{7}{l}{$\ddag$ and $\dag$ indicate VARE performs significantly better than and equivalently to the corresponding algorithm, respectively.}\\
    \multicolumn{7}{l}{* The overall rank is computed by sorting the average rank values (shown in brackets) of five algorithms w.r.t. MIGD and MHV.}\\
    \end{tabular}%
  \label{tab:igd_hv}%
  \vspace{-5mm}
\end{table*}%
%\setlength{\textfloatsep}{0pt}

% Figure 4
\begin{figure}
    \centering
    \includegraphics[trim=70 25 5 10, clip, width=0.9\linewidth]{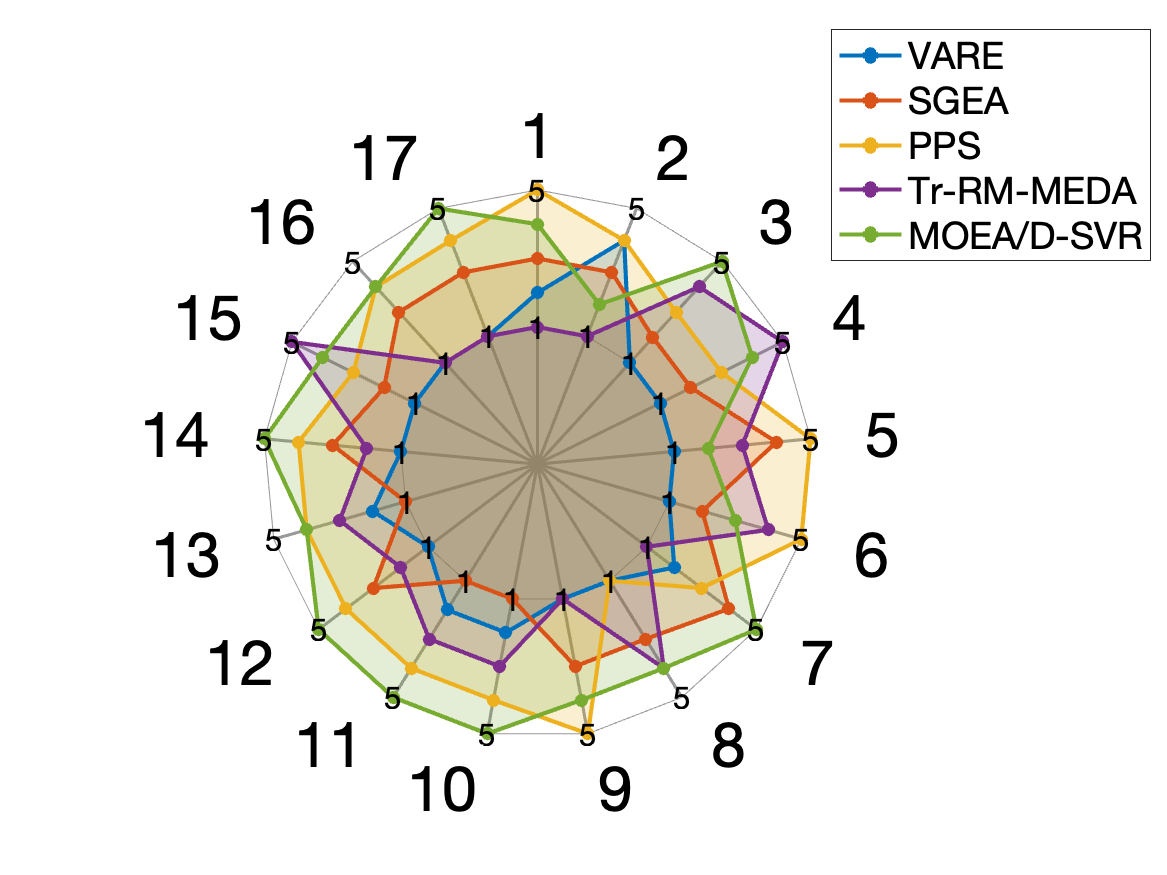}
    \caption{Spider plot of rank values of five algorithms w.r.t. MIGD and MHV, with problem No. as axis labels.}
    \label{fig:alg-rank}
\end{figure}

\section{Results} \label{sect:alg-cmp}
This section presents empirical results to illustrate that the proposed algorithm is generally more effective in solving the problems considered in this paper than the other compared algorithms.

% \subsection{Compared Algorithms}
Table \ref{tab:igd_hv} presents the mean and standard deviation values of MIGD and MHV obtained by five algorithms for 17 problems (No. 1--9 are biobjective and No. 10--17 are triobjective problems) with $(n_t, \tau_t)=(10,10)$, where statistical significance was performed by Wilcoxon-sum test with Bonferroni correction at a significance of 0.05. The results on other settings of $(n_t, \tau_t)$ can be found in the supplementary material. To rank these algorithms, we applied nondominated sorting \cite{Deb2002NSGA2} to sort the paired values (MIGD, MHV) of five algorithms for each problem, resulting in at most five fronts with the algorithms in the 1st front being assigned rank 1. Algorithms in the same front will have the same rank value, and the ranking result is shown in Fig.~\ref{fig:alg-rank}. After that, the rank values of each algorithm are averaged over the 17 problems, and the average values of all the algorithms are further sorted in an ascending order to determine the overall rank of each algorithm (see the last row of Table~\ref{tab:igd_hv}). The two measures are consistent in most of the cases with a few exceptions, which have been also observed in other studies \cite{Jiang2019SDP}. 

DF1 and DF2 are two relatively simple DMOPs where all the PS segments move in the same direction after a change, and DF2 involves the change of position-related variables which could cause drastic loss of population diversity. For these two problems, Tr-RM-MEDA performs best, followed by VARE for DF1 and MOEA/D-SVR for DF2. PPS's MIGD values are one order of magnitude bigger than the other algorithms, implying that it is ineffective to capture the movement of PS for these two problems. The relatively large IGD value by VARE for DF2 may suggest VARE does not handle diversity loss as well as Tr-RM-MEDA and MOEA/D-SVR.

%%% Figure 19 %%%%%
\begin{figure*}[t]
	\centering
	\begin{subfigure}{.4\linewidth}
		\centering
		\includegraphics[width=0.99\linewidth]{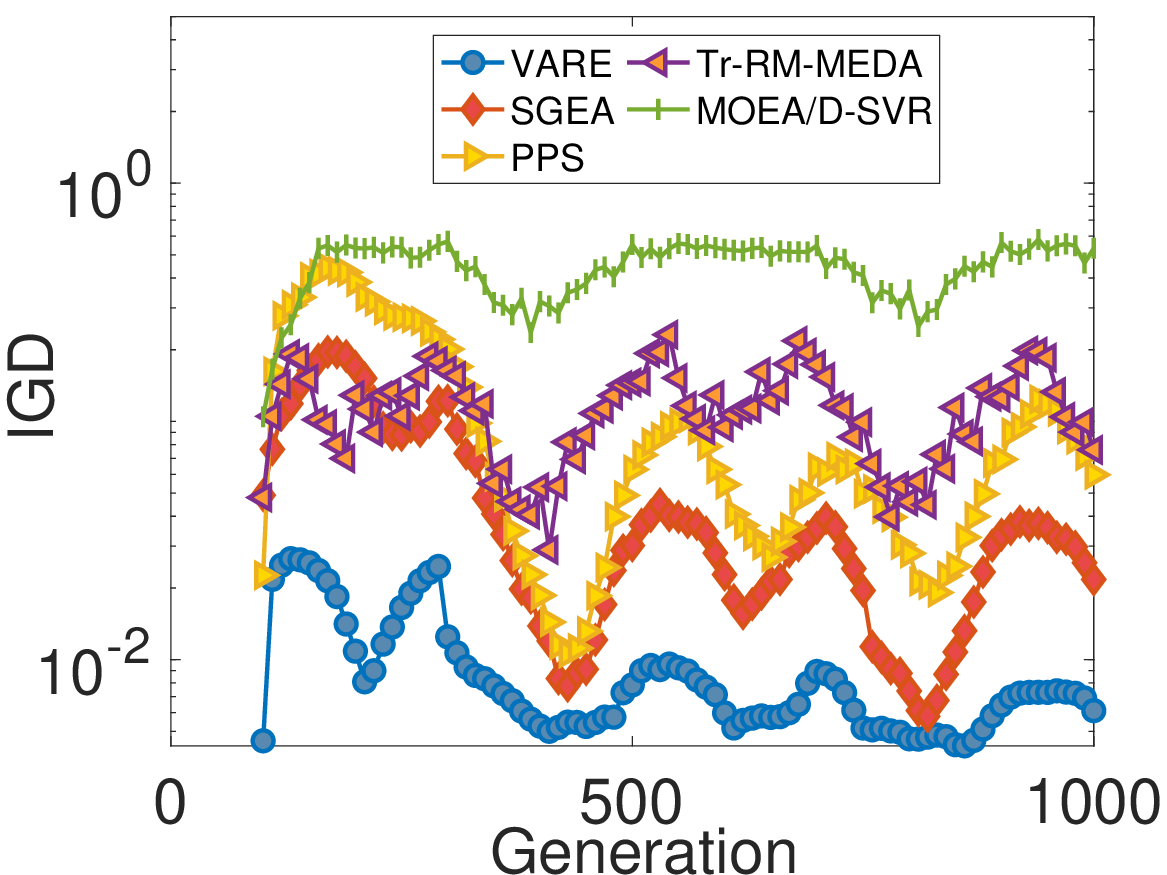}
		\caption{DF3}
		\vspace{-1mm}
	\end{subfigure}%
	\begin{subfigure}{.4\linewidth}
		\centering
		\includegraphics[width=0.99\linewidth]{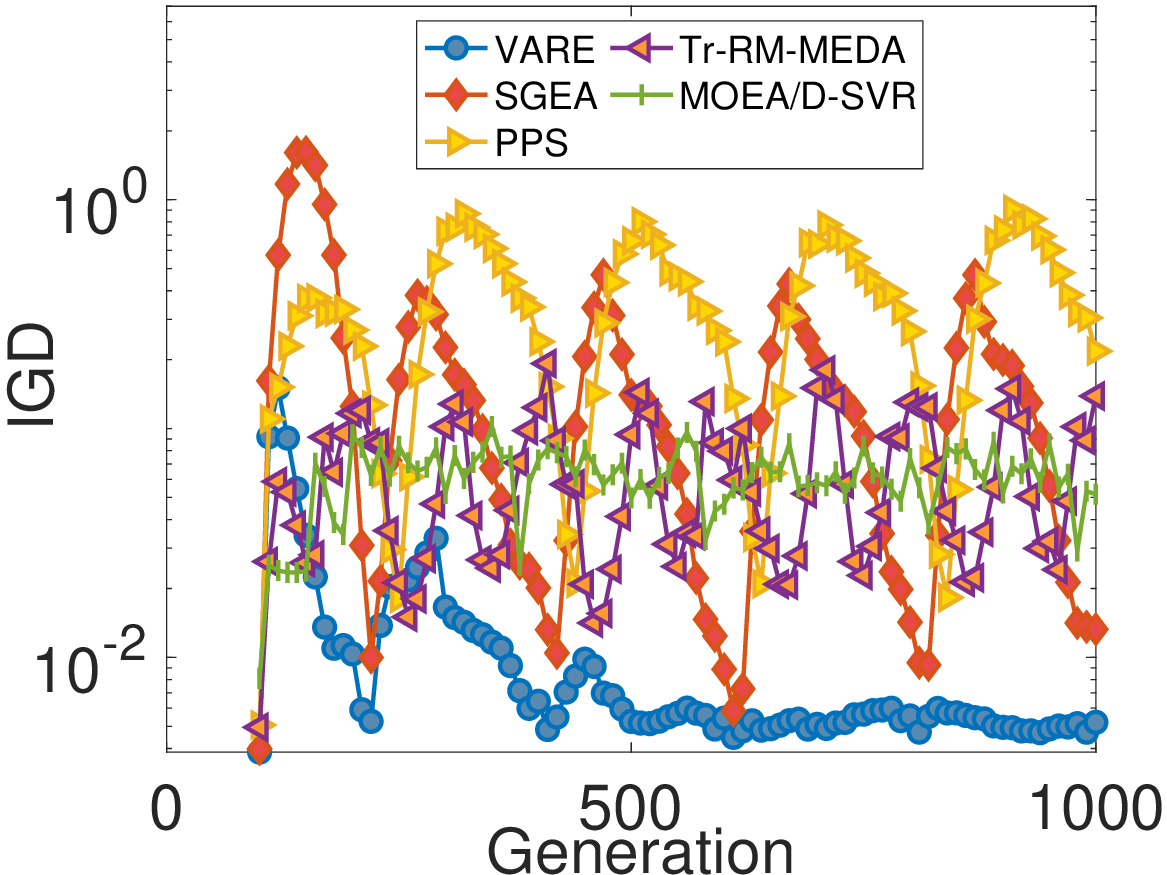}
		\caption{DF5}
		\vspace{-1mm}
	\end{subfigure} \\%[-2mm]
        %\hspace{-1.5mm}
	\begin{subfigure}{.4\linewidth}
		\centering
		\includegraphics[width=0.99\linewidth]{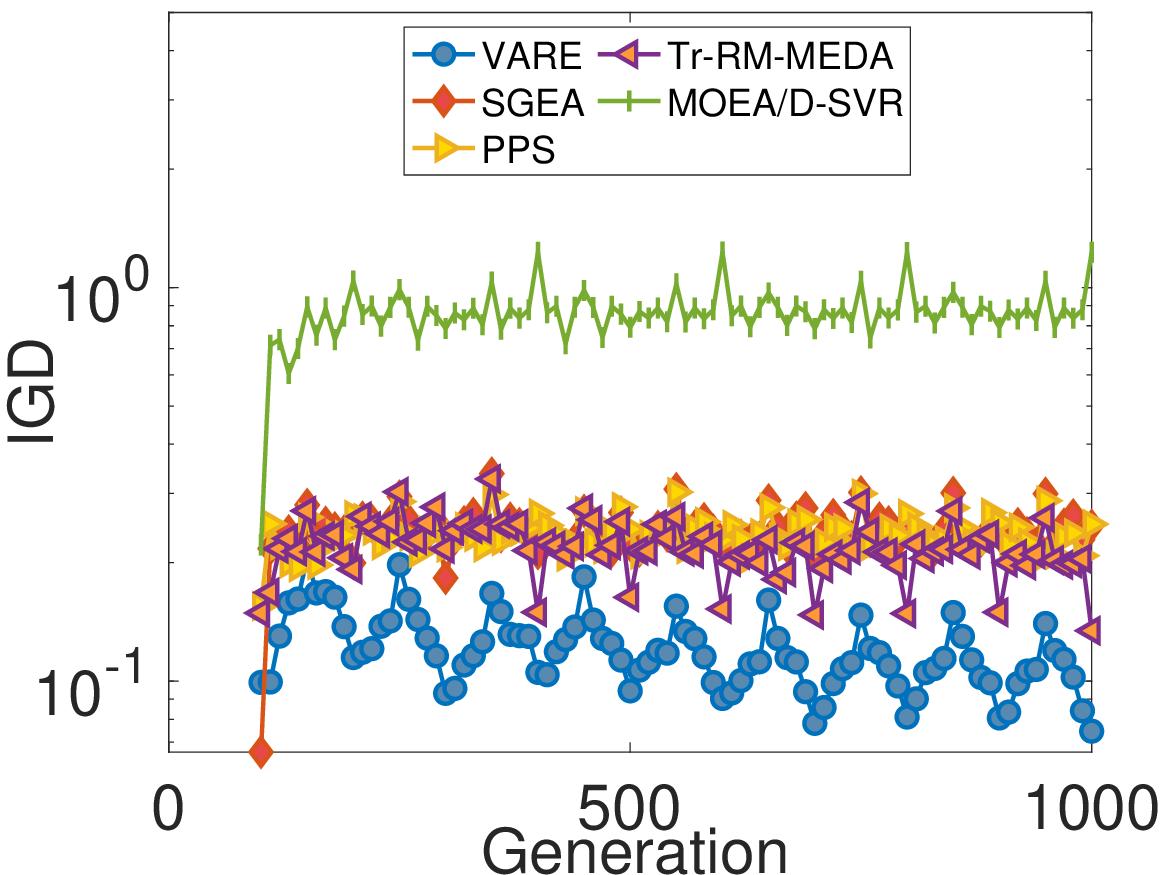}
		\caption{DF12}
		\vspace{-1mm}
	\end{subfigure}%
	\begin{subfigure}{.4\linewidth}
		\centering
		\includegraphics[width=0.99\linewidth]{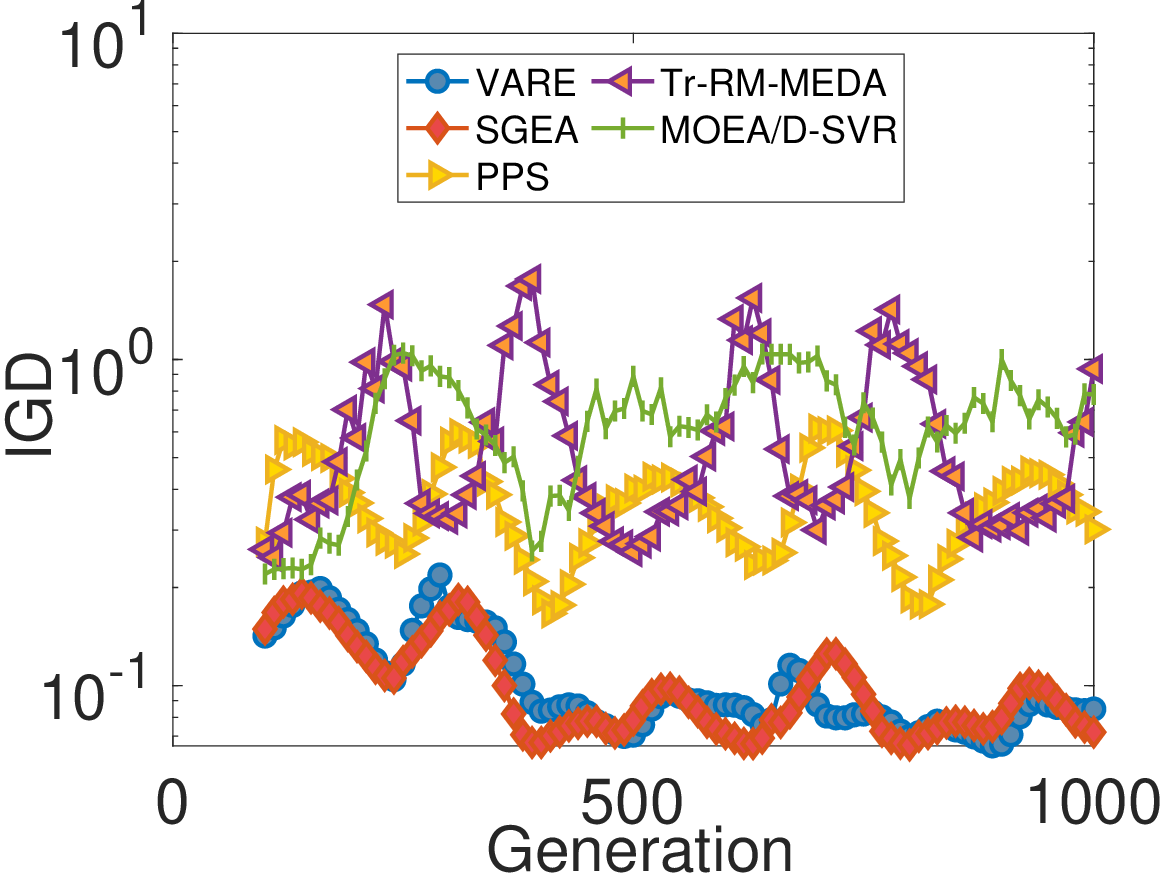}
		\caption{F8}
		\vspace{-1mm}
	\end{subfigure}\\[-1mm]
	\caption{Evolution curve of the average IGD obtained by five algorithms on four test problems.}
	\label{fig:alg-igdc}
	\vspace{-3mm}
\end{figure*}
%\vspace*{-5mm}

DF3 and DF4 two DMOPs with explicit linkages between variables. Unsurprisingly, VARE performs best on both problems due to its vector autoregression models that have considered variable correlations. On the contrary, without consideration of variable linkages, PPS and MOEA/D-SVR have poor MIGD values on DF3. Tr-RM-MEDA performs badly on both problems, implying that its domain adaptation techniques struggle to address variable linkages. SGEA ranks second in such problems, probably due to its fast convergence enabled by steady-state evolution.

DF5 and DF6 have a changing convexity-concavity in PF geometries, and DF6 additionally has multiple local optima. For such types of problems, VARE is more effective than the other algorithms, although it also faces the same convergence issues as the others on DF6 as suggested by large MIGD values. SGEA is less effective than VARE, but much better than the remaining algorithms. The poor performance of all 5 five algorithms on DF6 suggests that the change from a global optimum to a local one is a challenging dynamic, especially in fast-changing environments.

The PS of DF7 and DF8 can move in different directions and with different stepsizes after a change, thereby challenging centroid-based prediction methods. It is clear that PPS and SGEA, both relying on population centroids for change handling, are significantly worse than the multi-directional prediction methods VARE and MOEA/D-SVR for DF7. Tr-RM-MEDA is the best performer on DF7 due to the use of centroid-free domain adaptation techniques. However, it has poor performance on DF8 as DF8 has multiple knees on the PF which makes domain adaptation ineffective. DF9 is a PF-disconnected DMOP, for which VARE and Tr-RM-MEDA perform equally well. PPS and MOEA/D-SVR seem struggling to solve such problems.

The rest of 8 problems (No. 10-17) are triobjective problems with more or less similar dynamics to the above 9 biobjective ones except for DF14, which has a degenerate PF over time. VARE is slightly inferior to SGEA on DF10, DF11 and DF13, but it generally is the best performer on the other remaining 3-objective problems, demonstrating that VARE is still effective in addressing dynamic changes with more objectives. PPS and MOEA/D-SVR are the bottom performers for these 3-objective problems, and Tr-RM-MEDA overall is the middle performer in this case.

It is observed from Fig.~\ref{fig:alg-rank} that VARE is a top performer for almost all the problems except DF2, on which its two indicator values are still close to those of the other algorithms. SGEA and Tr-RM-MEDA are middle performers, followed by PPS and MOEA/D-SVR as bottom performers that are effective only for a couple of problems. The overall rank in the last row of Table \ref{tab:igd_hv} further confirms that VARE is the best and MOEA/D-SVR is the least effective algorithm for the considered problems.

%%%% Fig 3 %%%%
% {\setlength{\textfloatsep}{-10pt}} % Vertical space below (above) [t] ([b]) floats
% {\setlength{\abovecaptionskip}{-5pt}}
% {\setlength{\belowcaptionskip}{-5pt}}
\begin{figure*}[t!]
	\small
	\centering
	\vspace{-4mm}
        \setlength{\tabcolsep}{-2pt}
	\begin{tabular}{cccc}
		(a)~~ & \begin{tabular}{ccc}
			\includegraphics[trim=0 0 40 20, clip, trim=0 0 40 20, clip, width=6cm,height=4cm]{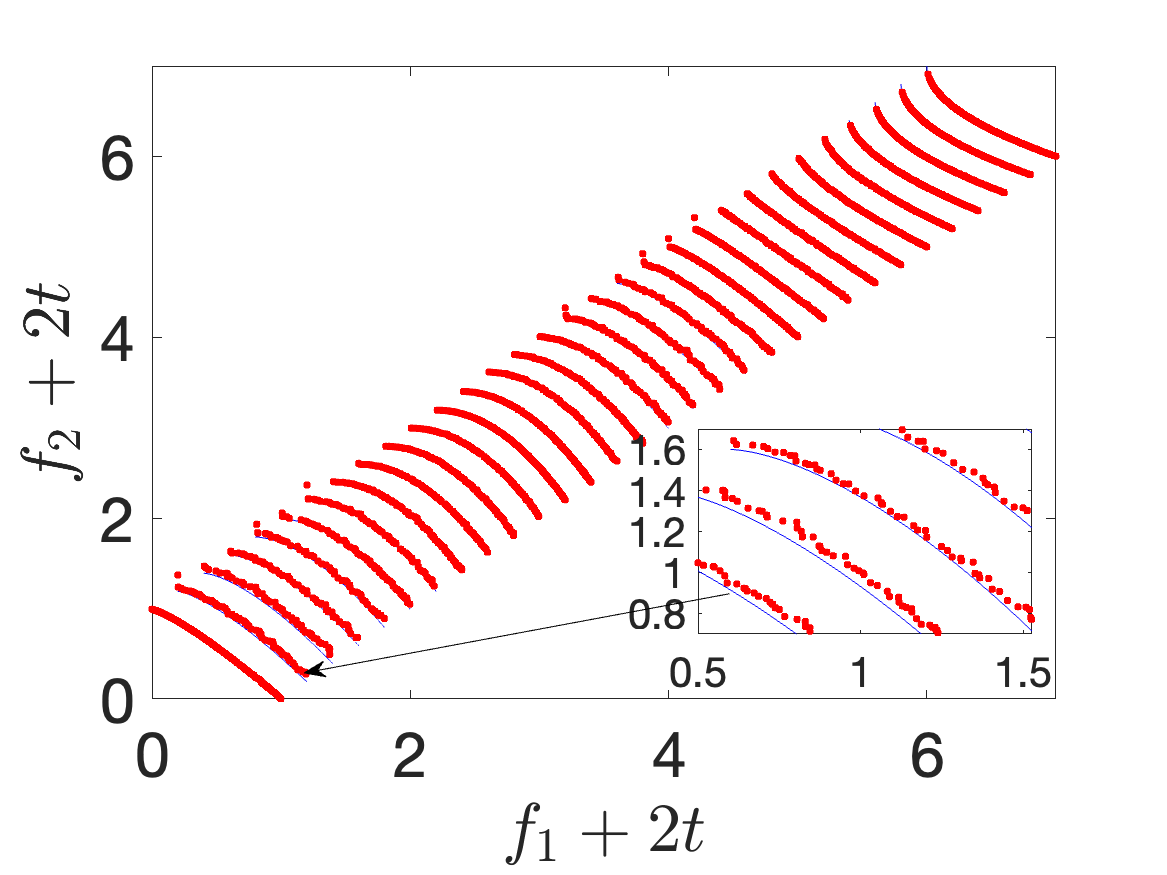} &
			\includegraphics[trim=0 0 40 20, clip, width=6cm,height=4cm]{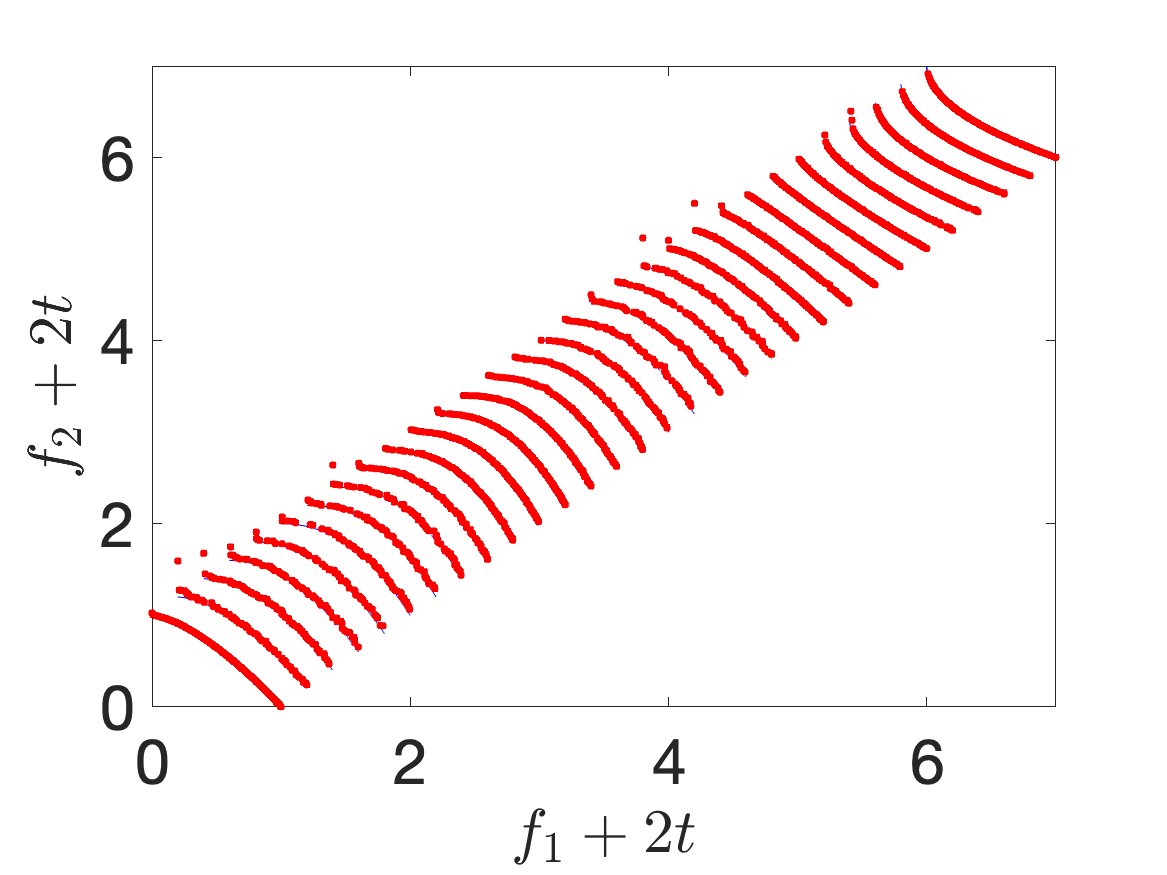} &
			\includegraphics[trim=0 0 40 20, clip, width=6cm,height=4cm]{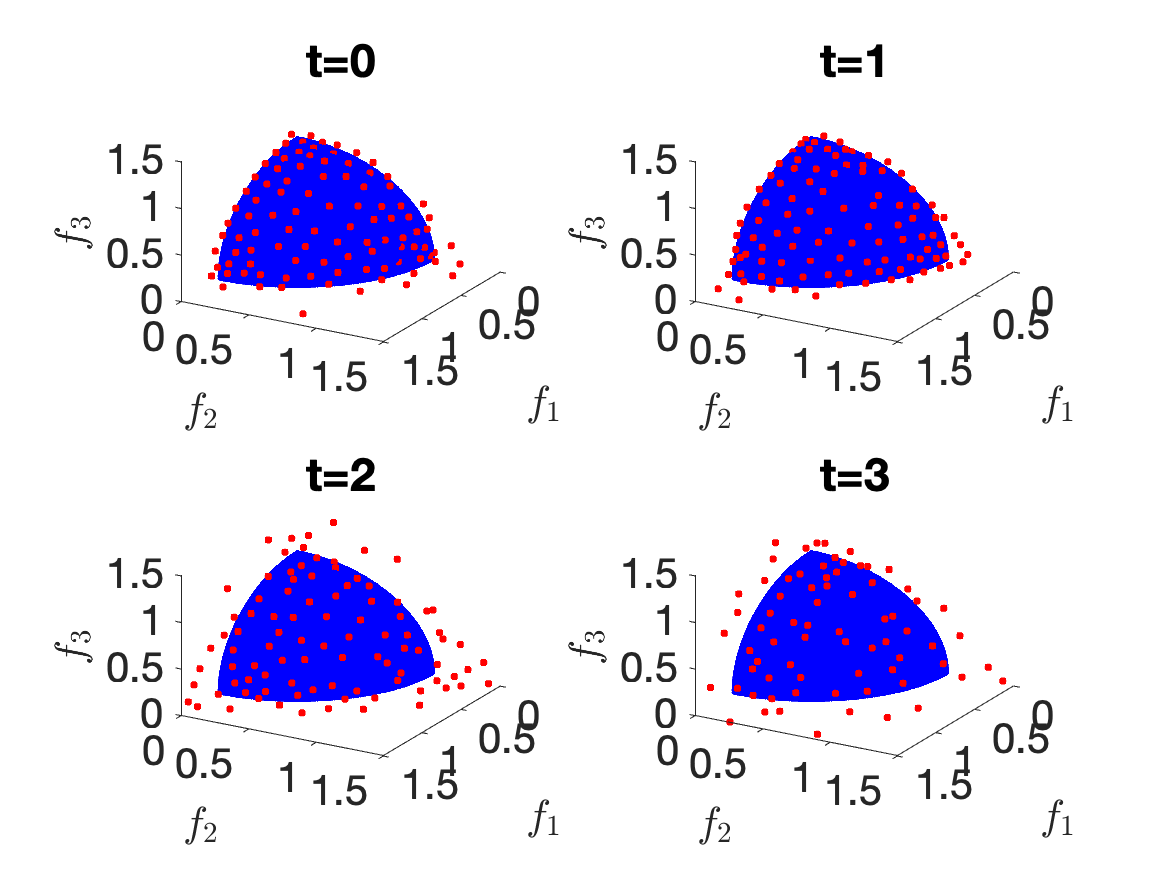} 
		\end{tabular}\\[-1.5mm]
		(b)~~ & \begin{tabular}{ccc}
			\includegraphics[trim=0 0 40 20, clip, width=6cm,height=4cm]{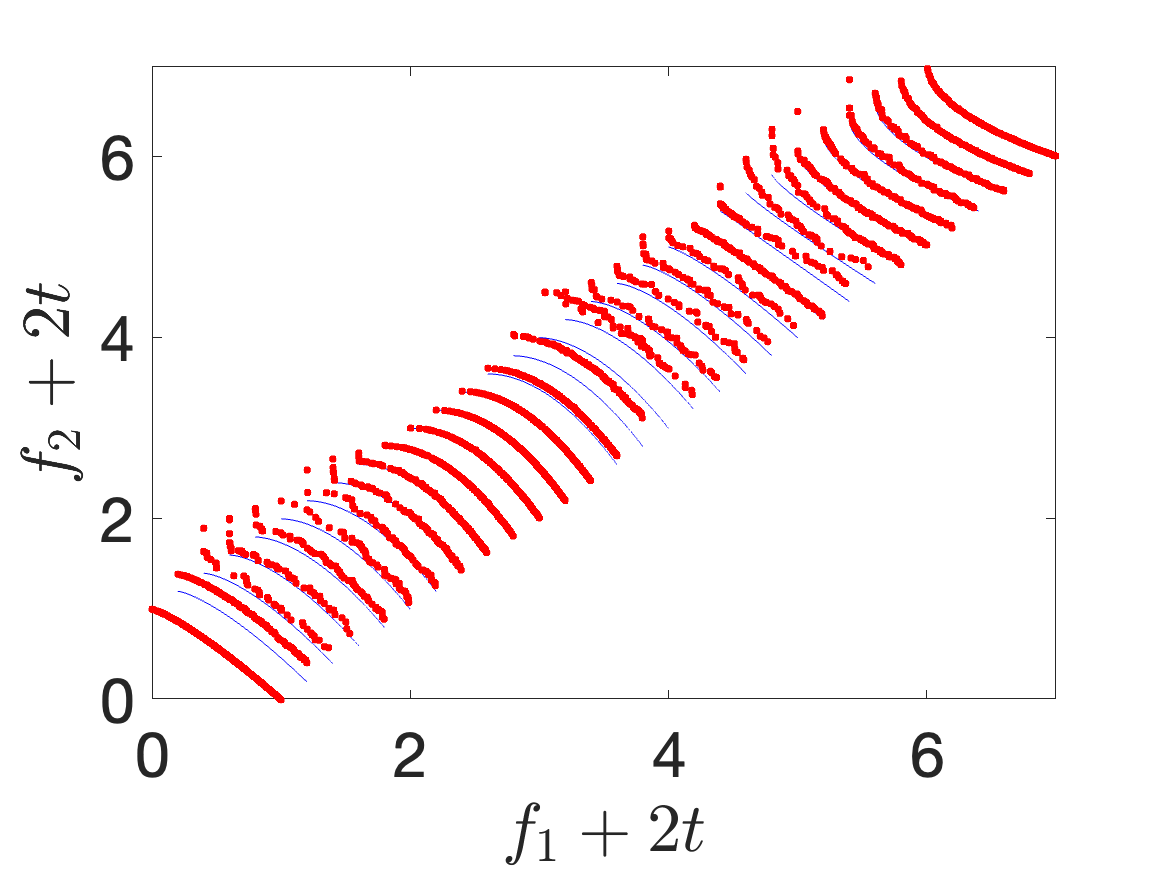} &
			\includegraphics[trim=0 0 40 20, clip, width=6cm,height=4cm]{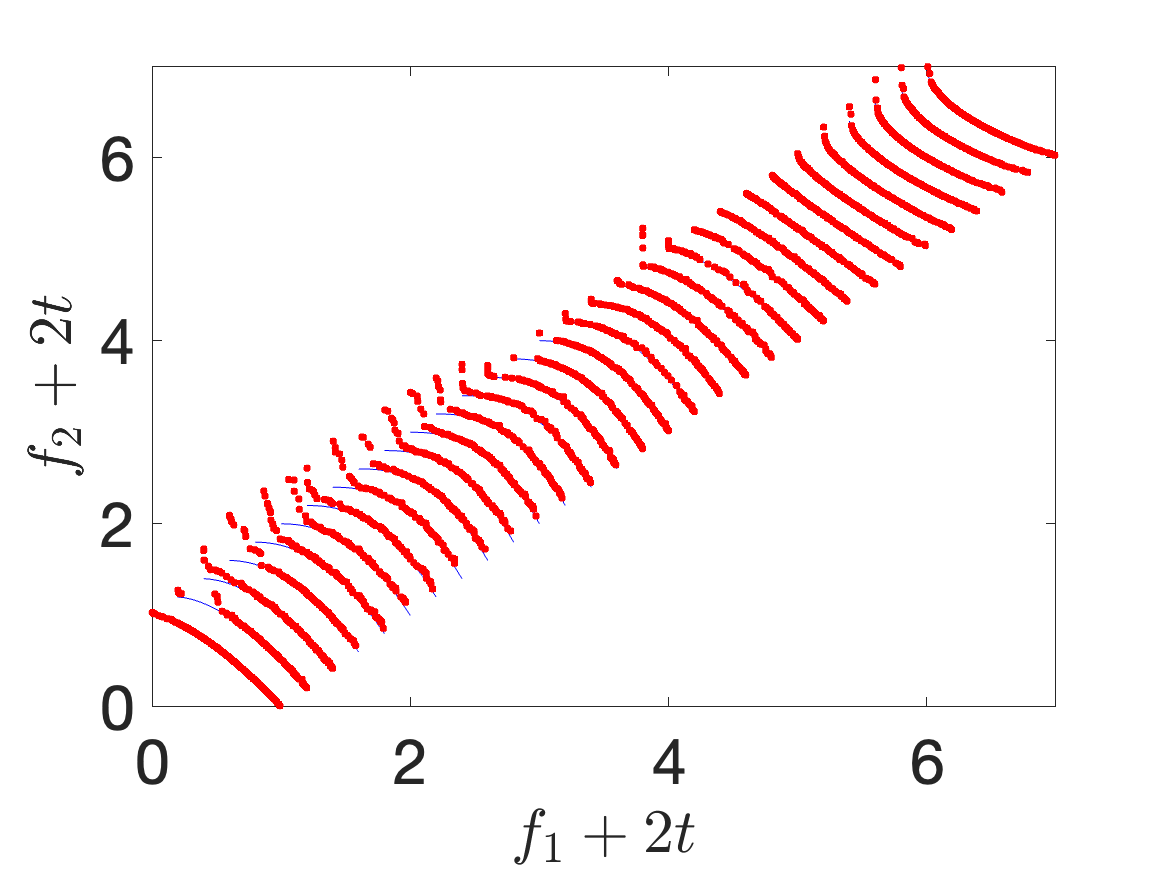} &
			\includegraphics[trim=0 0 40 20, clip, width=6cm,height=4cm]{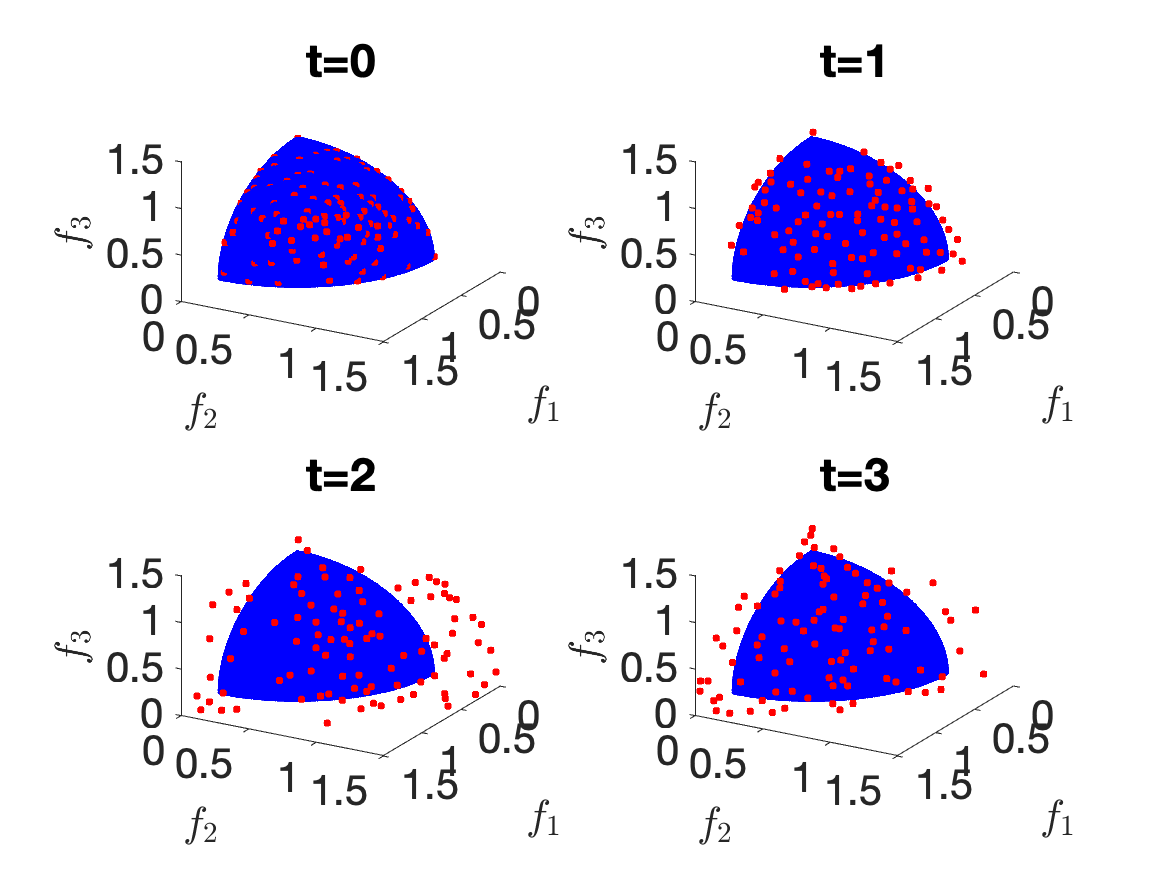} 
		\end{tabular}\\[-1.5mm]
		(c)~~ & \begin{tabular}{ccc}
			\includegraphics[trim=0 0 40 20, clip, trim=0 0 40 20, clip, width=6cm,height=4cm]{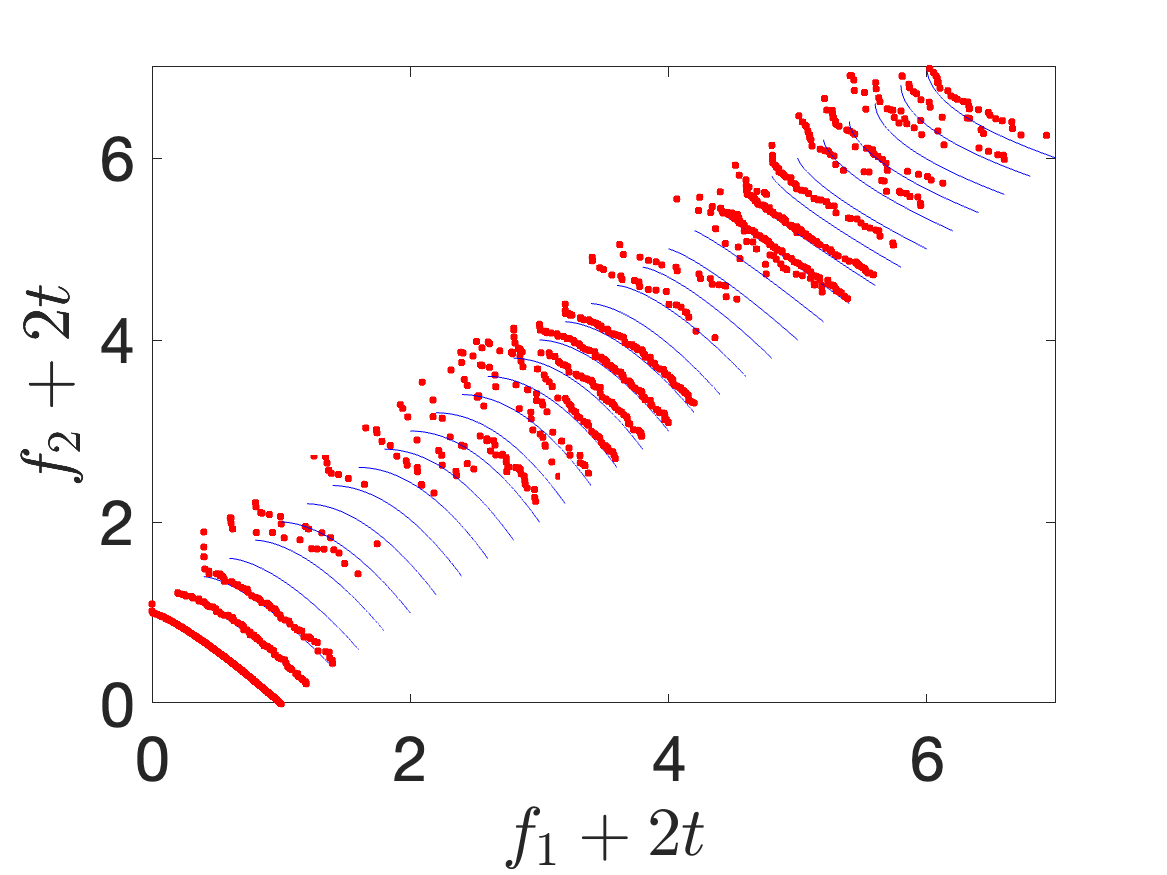} &
			\includegraphics[trim=0 0 40 20, clip, width=6cm,height=4cm]{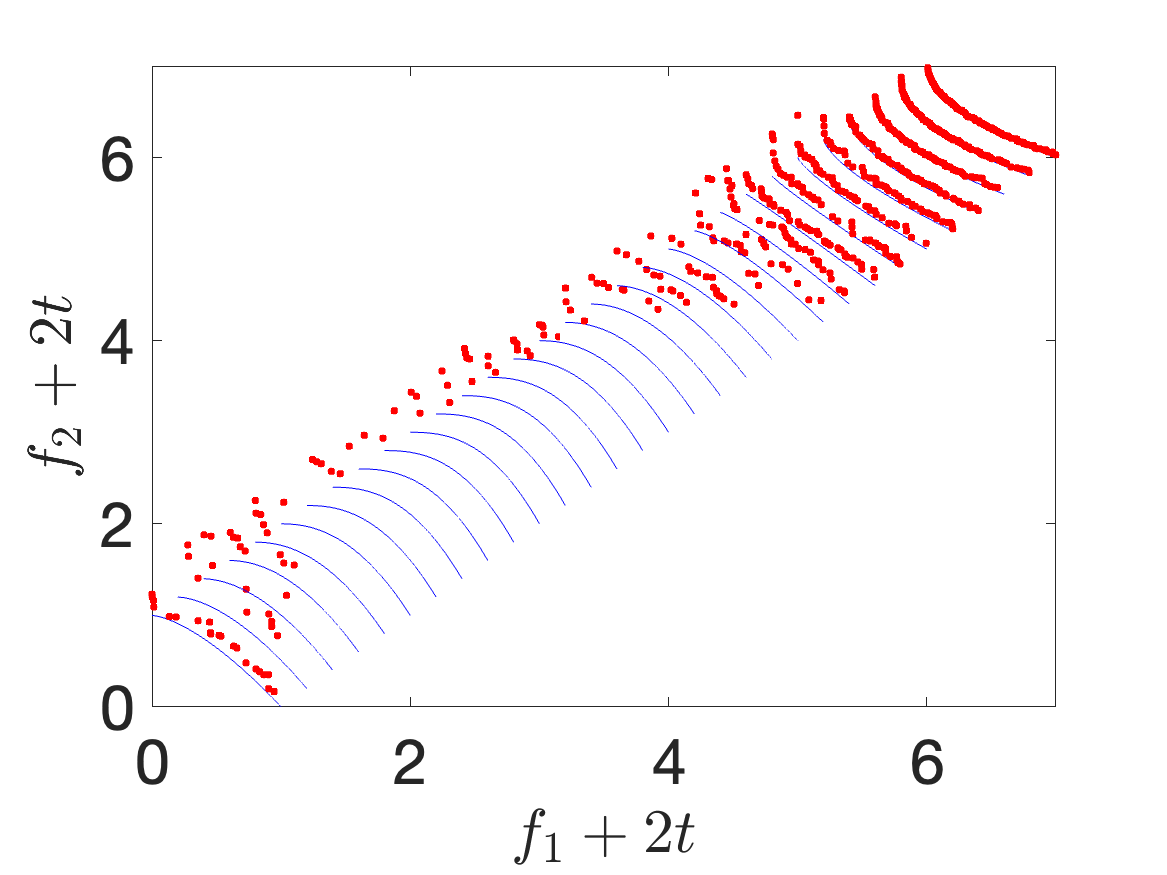} &
			\includegraphics[trim=0 0 40 20, clip, width=6cm,height=4cm]{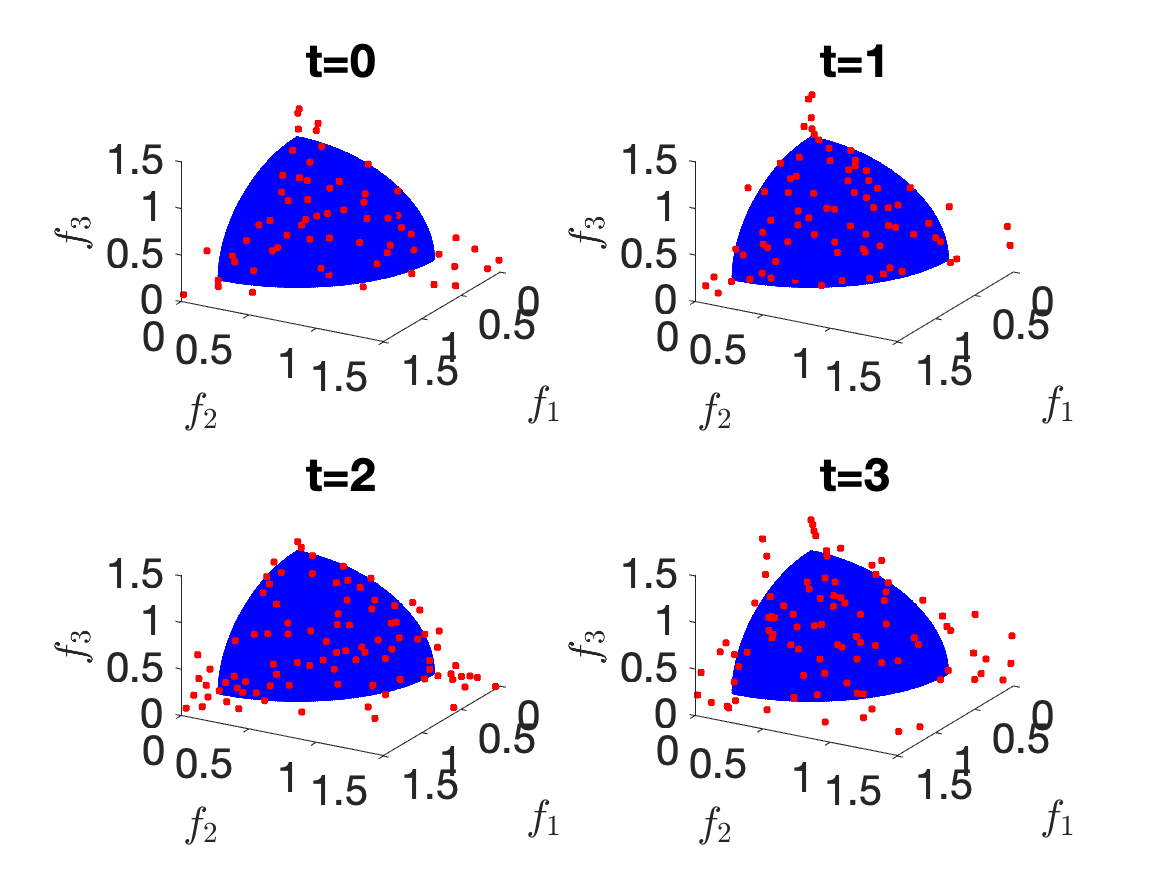}  
		\end{tabular}\\[-1.5mm]
		(d)~~ & \begin{tabular}{ccc}
			\includegraphics[trim=0 0 40 20, clip, width=6cm,height=4cm]{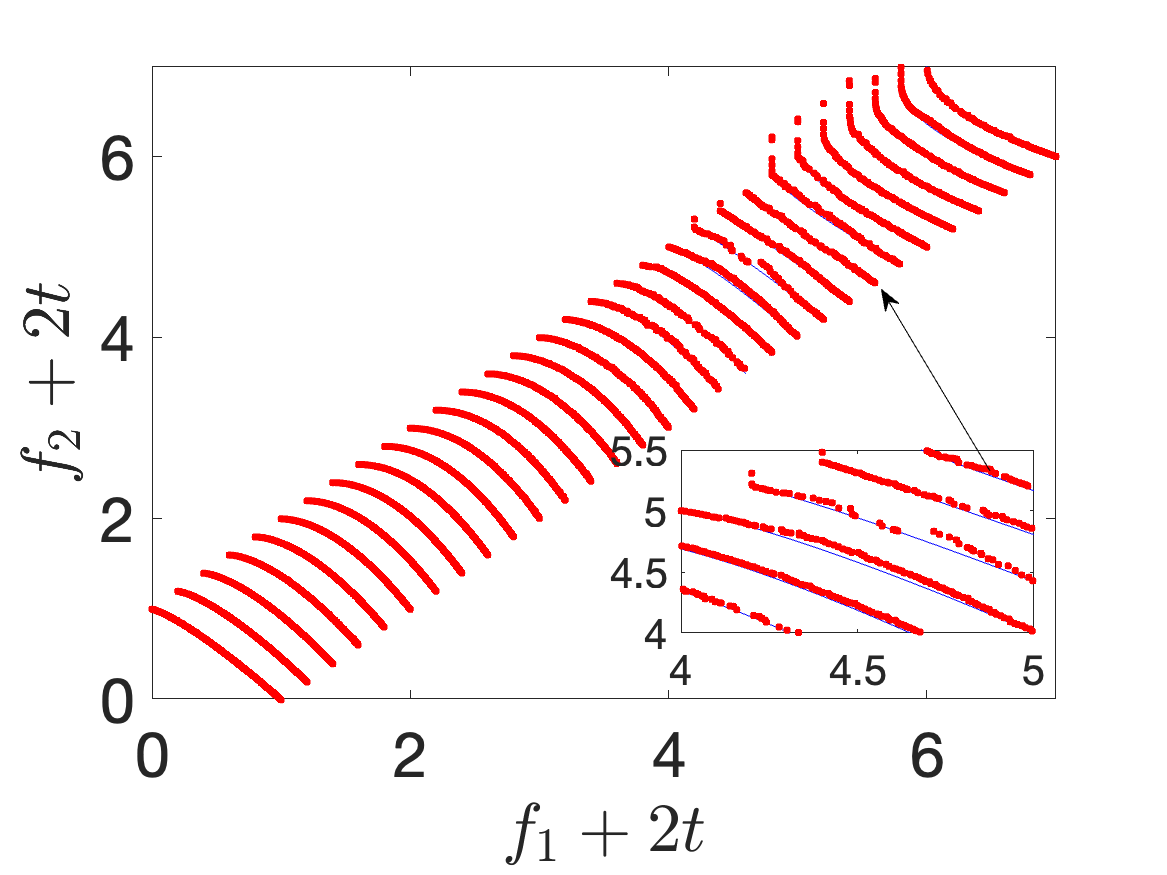} &
			\includegraphics[trim=0 0 40 20, clip, width=6cm,height=4cm]{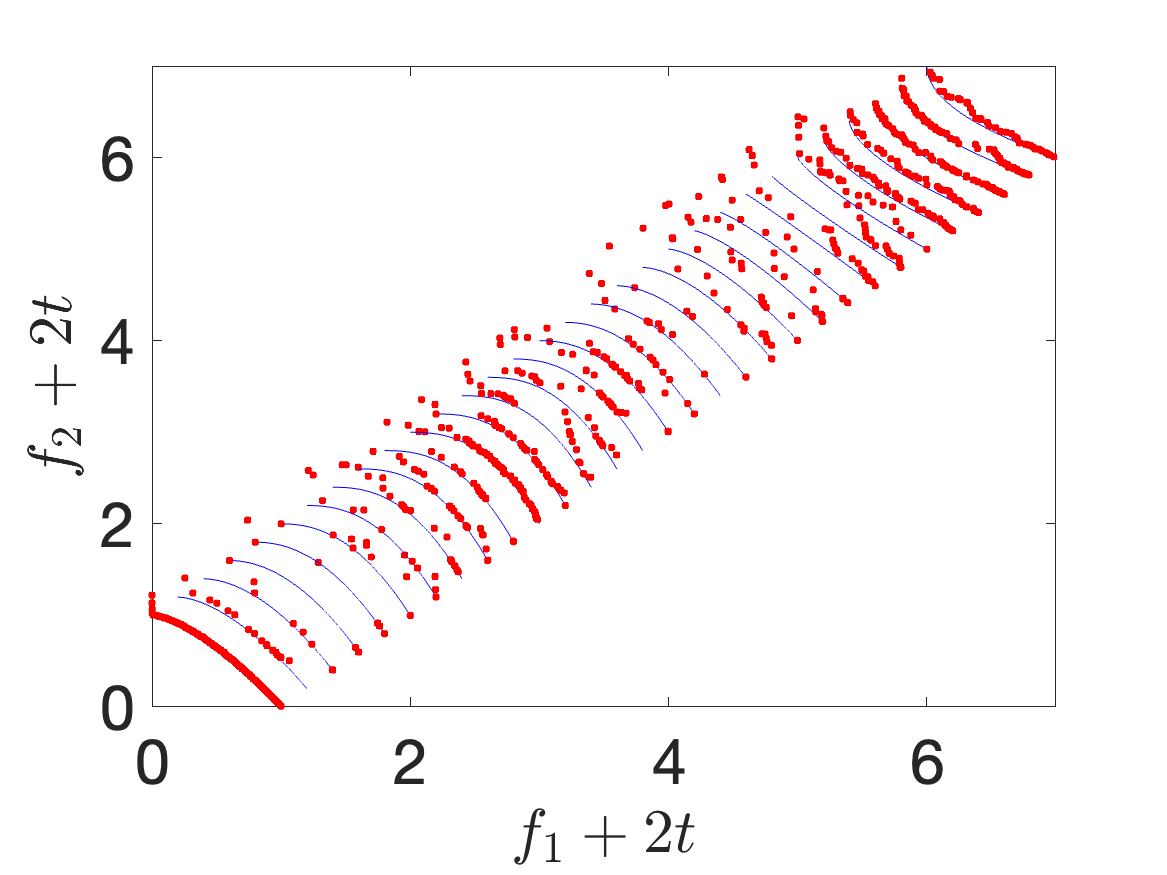} &
			\includegraphics[trim=0 0 40 20, clip, width=6cm,height=4cm]{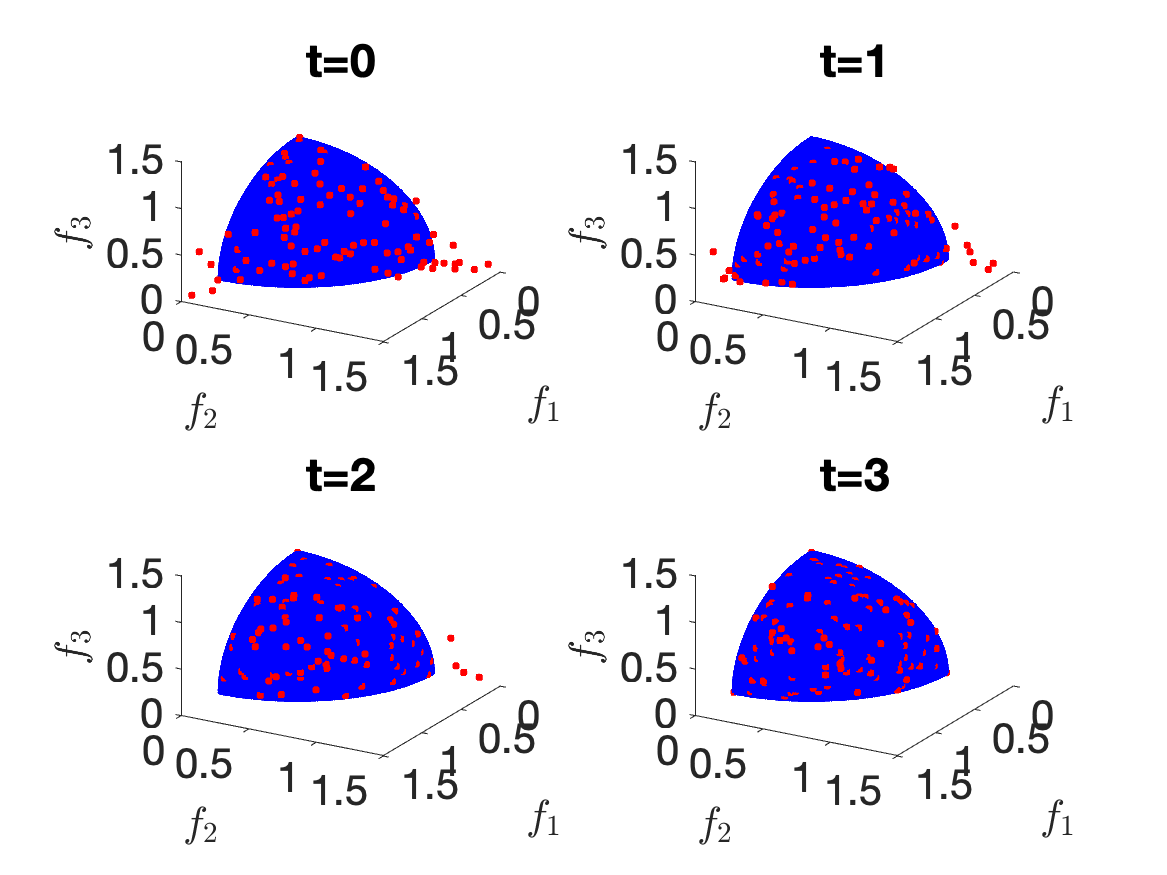} 
		\end{tabular}\\[-1.5mm]
		(e)~~ & \begin{tabular}{ccc}
			\includegraphics[trim=0 0 40 20, clip, width=6cm,height=4cm]{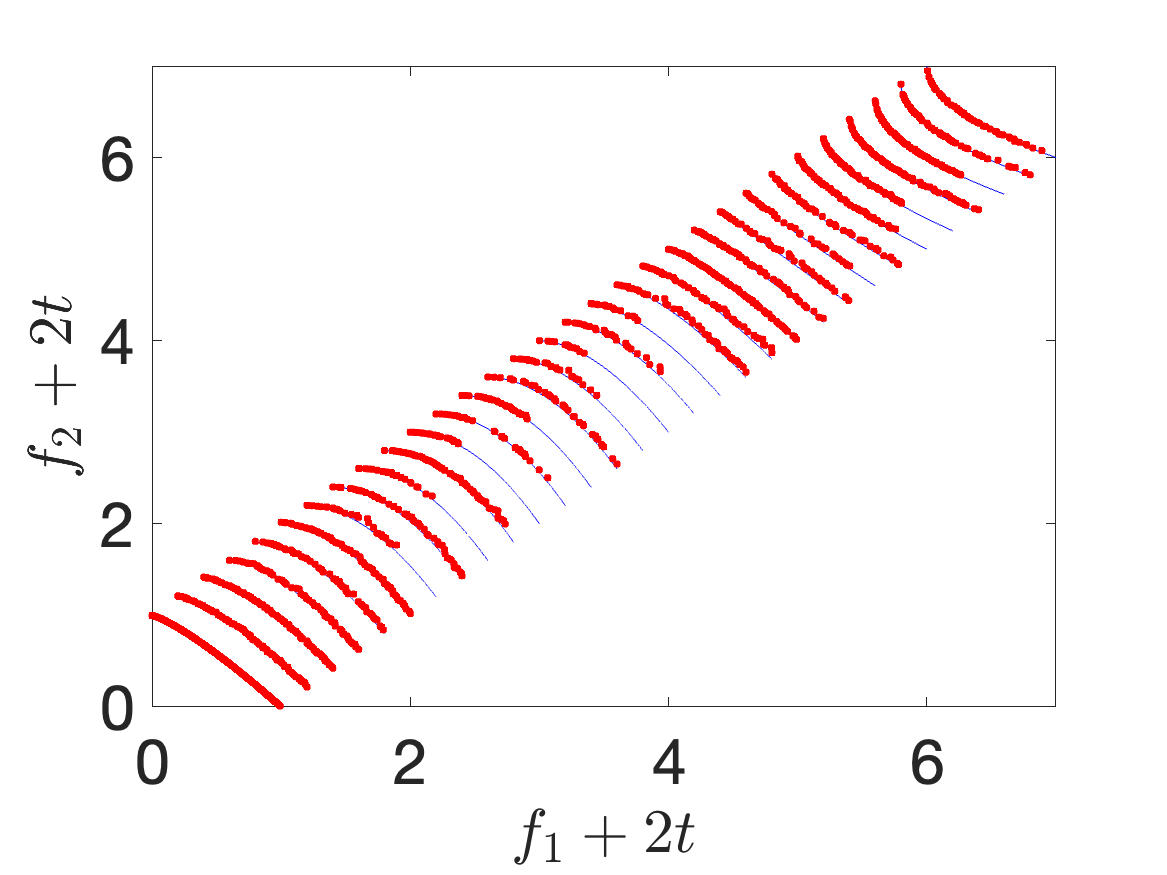} &
			\includegraphics[trim=0 0 40 20, clip, width=6cm,height=4cm]{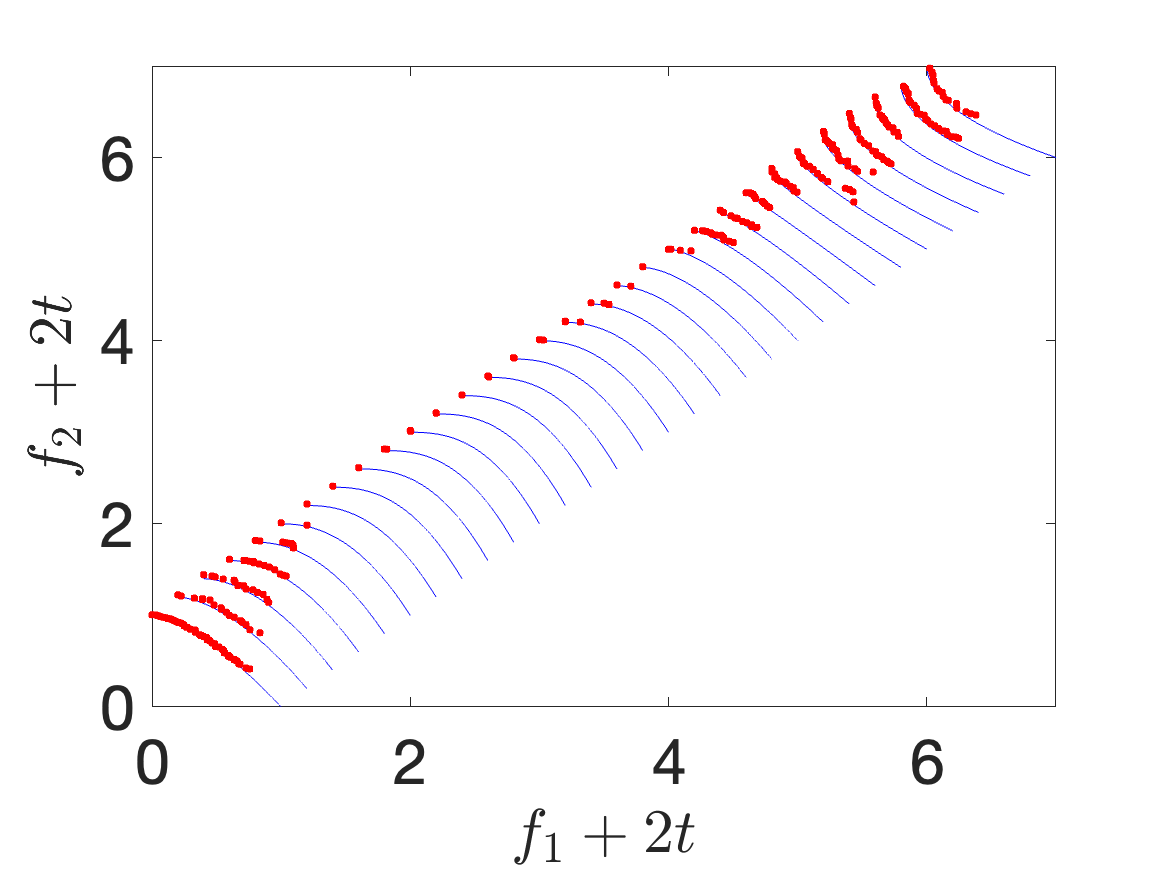} &
			\includegraphics[trim=0 0 40 20, clip, width=6cm,height=4cm]{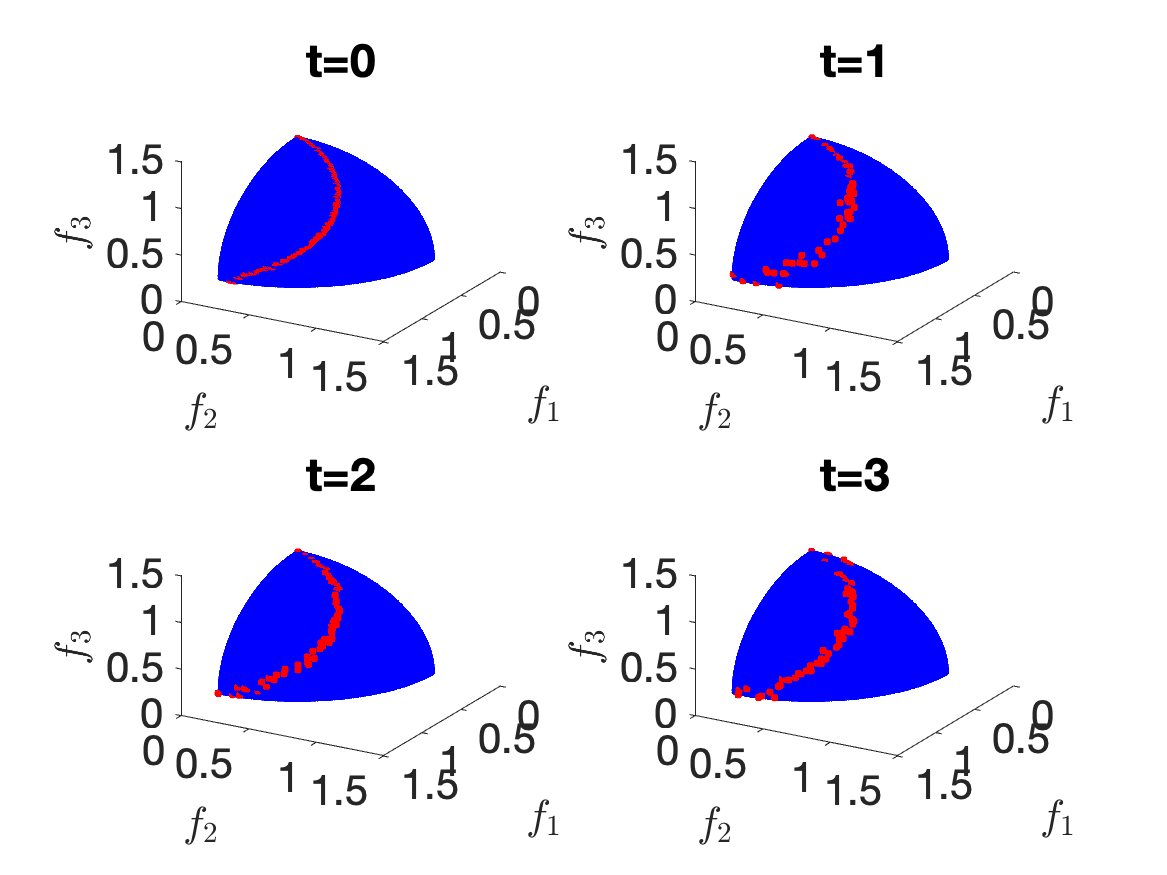} 
		\end{tabular}\\[-1.5mm]
	\end{tabular}\\[-2mm]
	\caption{PF approximations for DF1 (left column), DF3 (middle) at first 30 changes and FDA4 (right) at 4 specific changes by algorithms (a) VARE, (b) SGEA, (c) PPS, (d) Tr-RM-MEDA and (e) MOEA/D-SVR.}
	\label{fig:PA-1-4-FDA4}
	\vspace{-6mm}
\end{figure*}
%\vspace{-4mm}

We further investigate the evolutionary processes of different algorithms through displaying the IGD value of each aglorithm right before each change. Fig.~\ref{fig:alg-igdc} shows VARE is able to reach IGD values that are much lower than or no worse than the other algorithms for the selected problems. MOEA/D-SVR performs poorly although it has the least fluctuation in its IGD values. The remaining three algorithms achieve low IGD values in some generations but mostly large values throughout the evolutionary process. Fig. \ref{fig:PA-1-4-FDA4} presents the PF approximations of five algorithms for a few selected problems. For the two biobjective problems, VARE tracks the changing PF effectively. SGEA tracks most of the changes well but has issues in population distribution or convergence for some changes. PPS loses population diversity but seems to become better as more changes occur. This can be explained by the fact that PPS requires long time-series data to build a effective population prediction model. Tr-RM-MEDA tracks the PF well for DF1 but experiences loss of population diversity for DF3. MOEA/D-SVR faces diversity loss in both problems, although this happens less frequently in DF1. For the 3-objective FDA4, all the algorithms except MOEA/D-SVR face convergence issues. 

VARE has the best population distribution across the PF. MOEA/D-SVR has the biggest loss of population diversity, although its population is well converged in each of the four considered changes. This demonstrates the adoption of a diversity-focused strategy as done in VARE is necessary for addressing dynamic changes.

\begin{figure}
    \centering
    \includegraphics[width=0.8\linewidth]{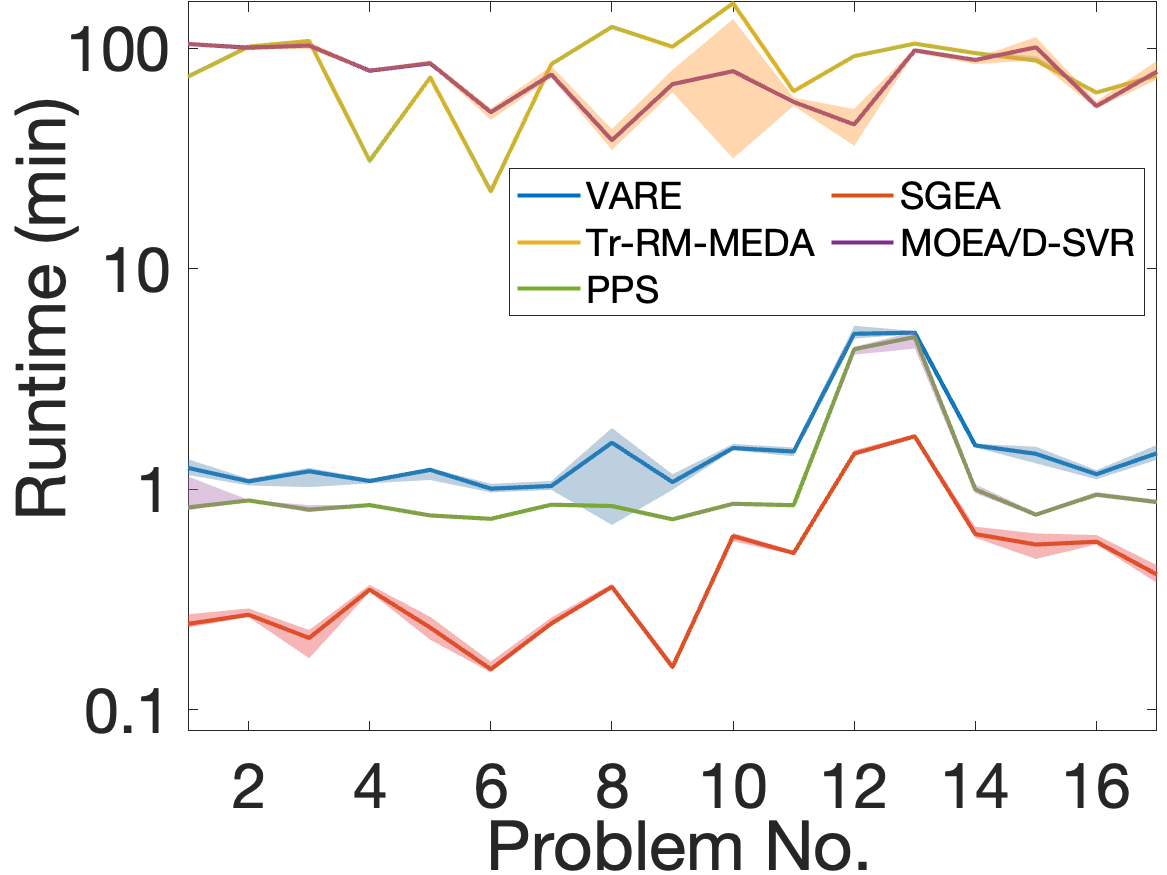}
    \caption{Average execution time of five algorithms for a single run of each problem, with shaded region bounded by max and min values.}
    \label{fig:runtime}
\end{figure}

In addition, runtime is another focus of the proposed algorithm. It is observed from Fig.~\ref{fig:runtime} that SGEA is the fastest because it uses a simple and efficient prediction approach, followed by PPS and then VARE. These three algorithms are able to solve most of the problems, each with 100 changes, in less than 2 mins for a single run. This demonstrates the efficiency of the modelling techniques (including vector autoregression and dimensionality reduction) used in VARE. In comparison, both Tr-RM-MEDA and MOEA/D-SVR are computationally intensive, requiring 100 mins in each run (2.1 days for 30 independent runs). Indeed, it took us 36 days for each of them to obtain the results presented in Table~\ref{tab:igd_hv}.
% It is also noticed that the three efficient algorithms consume higher execution time for DF12-13 than for the other problems. 

\section{Discussion}
The section is devoted to the analysis of key components of the proposed algorithm including parameter sensitivity.

\subsection{Component Analysis}
VARE consists of two key components, i.e., VAR and EAH. Here we would like to understand the role each of them plays in helping VARE to achieve promising results as shown in the previous section. Thus, we run VARE, VAR (by deactivating EAH in VARE) and EAH (by deactivating VAR in VARE) independently on the 17 test problems with $(n_t, \tau_t)=(10,10)$.

\begin{figure}
    \centering
    \begin{tabular}{cc}
    \includegraphics[width=0.48\linewidth, trim= 60 0 15 10,clip]{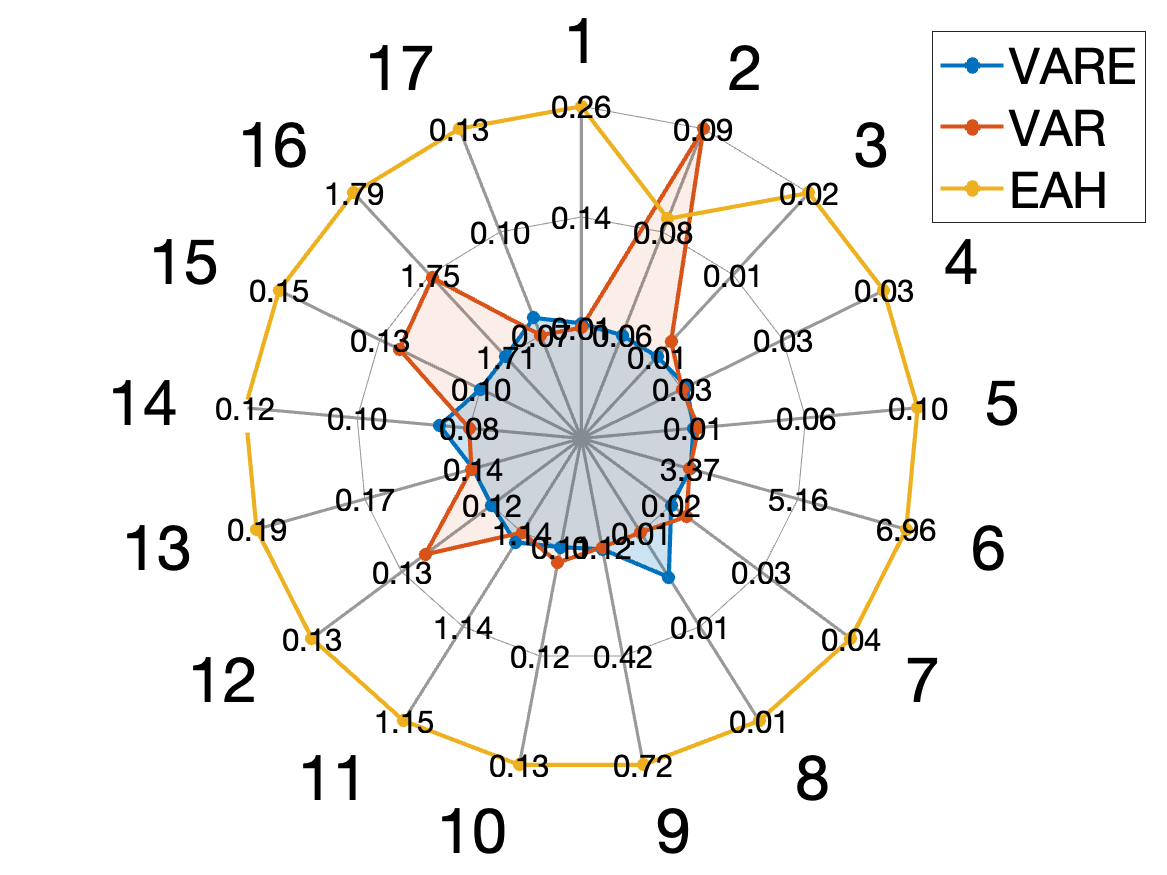}&
    \includegraphics[width=0.48\linewidth, trim= 65 0 10 10,clip]{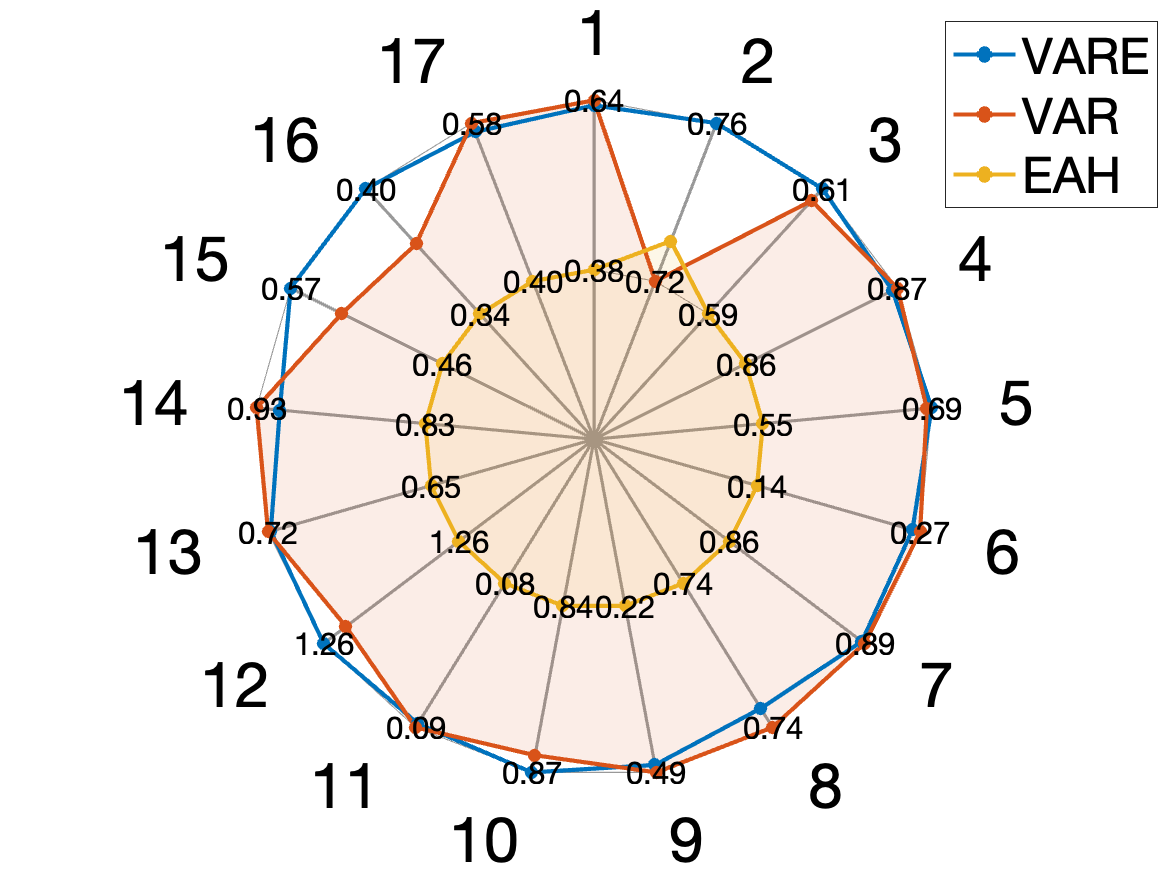}\\[-1mm]
    (a) MIGD & (b) MHV\\[-2mm]
    \end{tabular}	
    \caption{Spider plot of average MIGD values obtained by different algorithms with problem No. as axis labels.}
    \label{fig:variants}
    %vspace{-3mm}
\end{figure}

It can be observed from Fig.~\ref{fig:variants} that VAR is as effective as VARE for most of the problems except DF2 (No.2), DF12 (No.12), F8 (No.15), FDA4 (No.16). This means VAR has a major contribution to the proposed algorithm. EAH is significantly worse than VARE and VAR, indicating reliance only on hypermuation is not adequate to solve any of the 17 test problems effectively. Despite that, EAH does improve the performance of VARE, especially for several above-mentioned problems where VAR struggles to address. Therefore, we can conclude that VAR contributes most to VARE, followed by EAH.

\subsection{Influence of Lag Order $l$}\label{subsect:lo}
The lag order $l$ in VARE is a key parameter that requires tuning, although there are methods in VAR literature to automatically determine the optimal $l$ from a purely statistical perspective for large datasets. However, in this paper, we consider $l$ as a tunable parameter and test its influence on VARE. $l \in \{1, 2, 5, 8, 10, 15\}$ was tested on some selected problems with $(n_t, \tau_t)=(10, 10)$.

\begin{figure}
    \centering
    \includegraphics[width=0.8\linewidth]{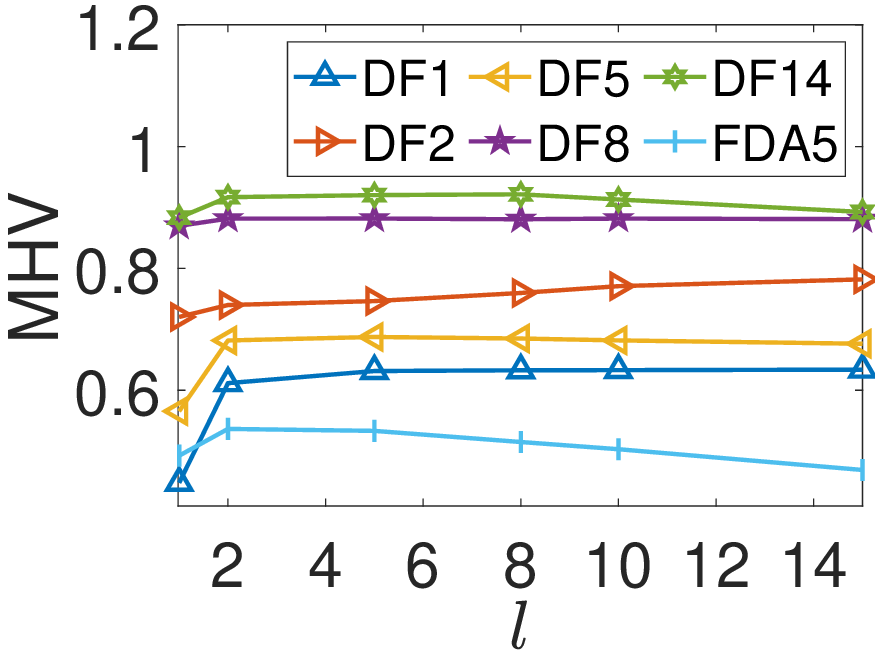}
    \caption{Average MHV values obtained by VARE with different lag order settings.}
    \label{fig:sens-p}
\end{figure}

Fig. \ref{fig:sens-p} illustrates that $l$ has an impact on the performance of VARE for the considered problems. Generally, VARE has growing performance as $p$ increases from 1, levels out between $l=5$ and $8$, and deteriorates for large $l$ values. Thus, it is clear that $l=5$ is the best choice.

\subsection{Influence of Adaptive Change Response}
As mentioned earlier, we hypothesise that the number $L$ of past environments used to define the probability of prediction-based change response is strongly coupled with the lag order $l$. Here, we examine different $L$ values that are multiples of $l$, i.e., $L = \gamma*l$ where $\gamma \in \{1, 2, 3, 4, 5\}$ to see if the use of more past environments helps to improve VARE further.

\begin{figure}
    \centering
    \includegraphics[width=0.8\linewidth]{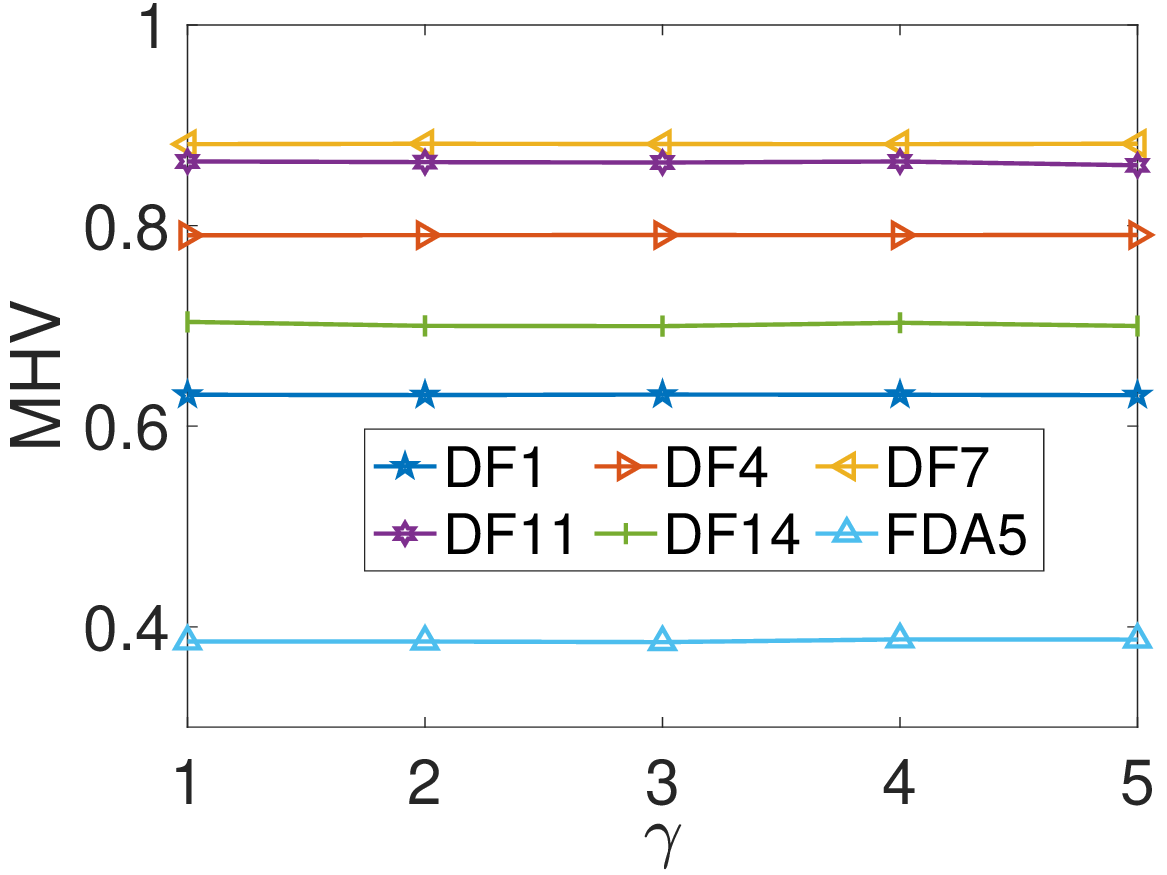}
    \caption{Average MHV values of VARE with different $\gamma$ settings on six problems with $(n_t, \tau_t)=(10, 10)$.}
    \label{fig:sens-gamma}
\end{figure}

It is observed from Fig.~\ref{fig:sens-gamma} that $L$ has little impact on VARE, which means considering more past environments is not helpful for determining the probability of prediction other than increasing computational calculations, although some larger $L$ values help VARE to obtain slightly better MHV values for problems such as DF1 and DF14. This is understandable since, when data has a trend, the autocorrelations for small lags tend to be significant because observations nearby in time are also nearby in size. Thus, it demonstrates that small lags are good enough for VARE to effectively determine sensible prediction probabilities. 

\section{Conclusion}
% Decomposition-based MOEAs are an important class of methods for multiobjective 
% optimization, and have been frequently shown to work well when proper scalarizing 
% functions are provided. In this paper, we have proposed two new scalarizing 
% functions which can induce controllable contours. By adjusting the size of 
% induced improvement regions, the new scalarizing functions can easily manage 
% population diversity. We have studied the influence of the new scalarizing 
% functions and have demonstrated that the proposed scalarizing functions with 
% proper improvement regions can significantly boost the performance of 
% decomposition-based MOEAs. Additionally, we have introduced an efficient 
% MOEA/D (i.e., eMOEA/D) framework based on the proposed scalarizing functions 
% and some new strategies. We have compared eMOEA/D with other recently-developed 
% approaches. The experimental results have clearly verified the effectiveness 
% of the eMOEA/D framework.

This paper has proposed for DMO a new EA, called VARE, that consists of two main components, i.e., vector autoregression (VAR) and environment-aware hypermutation (EAH), to address environmental changes effectively. VARE takes into account mutual relationship between variables, which has been neglected in existing DMO algorithms, when building predictive models to generate new solutions in new environments, enabling effective prediction of the new PF/PS after each environmental change. Additionally, VARE introduces EAH to rigorously handle dynamic environments in complement to VAR. EAH is used to address scenarios beyond the capability of VAR-based population prediction. With adaptive use of VAR and EAH, the proposed algorithm is able to solve DMOPs more effectively than other popular DMOEAs considered for comparison in this work while requiring only 2\% runtime of Tr-RM-MEDA and MOEA/SVR, as demonstrated by our empirical studies and analysis.

In this paper, VAR models are built independently for each reference direction at each new environment. Two refinements can be done in the future work: 1) building a parametric VAR model that can be tailored to different reference directions; and 2) online training/update of the VAR models for stream learning to reduce time complexity. In addition, it will be also interesting to investigate the performance of the proposed algorithm in high-dimensional objective space. 

\bibliographystyle{jabbrv_IEEEtran}
\bibliography{VARE}

\begin{thebibliography}{10}
\providecommand{\url}[1]{#1}
\csname url@samestyle\endcsname
\providecommand{\newblock}{\relax}
\providecommand{\bibinfo}[2]{#2}
\providecommand{\BIBentrySTDinterwordspacing}{\spaceskip=0pt\relax}
\providecommand{\BIBentryALTinterwordstretchfactor}{4}
\providecommand{\BIBentryALTinterwordspacing}{\spaceskip=\fontdimen2\font plus
\BIBentryALTinterwordstretchfactor\fontdimen3\font minus
  \fontdimen4\font\relax}
\providecommand{\BIBforeignlanguage}[2]{{%
\expandafter\ifx\csname l@#1\endcsname\relax
\typeout{** WARNING: IEEEtran.bst: No hyphenation pattern has been}%
\typeout{** loaded for the language `#1'. Using the pattern for}%
\typeout{** the default language instead.}%
\else
\language=\csname l@#1\endcsname
\fi
#2}}
\providecommand{\BIBdecl}{\relax}
\BIBdecl

\bibitem{Farina2004deb}
M.~Farina, K.~Deb, and P.~Amato, ``Dynamic multiobjective optimization
  problems: test cases, approximations, and applications,''
  \emph{\JournalTitle{IEEE Transactions on Evolutionary Computation}}, vol.~8,
  no.~5, pp. 425--442, 2004.

\bibitem{Jiang2022survey}
S.~Jiang, J.~Zou, S.~Yang, and X.~Yao, ``Evolutionary dynamic multi-objective
  optimisation: A survey,'' \emph{\JournalTitle{ACM Computing Surveys}},
  vol.~55, no.~4, pp. 1--47, 2022.

\bibitem{Eaton2017ACO}
J.~Eaton, S.~Yang, and M.~Gongora, ``Ant colony optimization for simulated
  dynamic multi-objective railway junction rescheduling,''
  \emph{\JournalTitle{IEEE Transactions on Intelligent Transportation
  Systems}}, vol.~18, no.~11, pp. 2980--2992, 2017.

\bibitem{Zhang2011AIS}
Z.~Zhang and S.~Qian, ``Artificial immune system in dynamic environments
  solving time-varying non-linear constrained multi-objective problems,''
  \emph{\JournalTitle{Soft Computing}}, vol.~15, no.~7, pp. 1333--1349, 2011.

\bibitem{huang2023large}
W.~Huang, H.~Ding, and J.~Qiao, ``Large-scale and knowledge-based dynamic
  multiobjective optimization for mswi process using adaptive competitive swarm
  optimization,'' \emph{\JournalTitle{IEEE Transactions on Systems, Man, and
  Cybernetics: Systems}}, 2023.

\bibitem{fu2022multiobjective}
J.~Fu, C.~Zou, M.~Zhang, X.~Lu, and Y.~Li, ``Multiobjective dynamic
  optimization of nonlinear systems with path constraints,''
  \emph{\JournalTitle{IEEE Transactions on Systems, Man, and Cybernetics:
  Systems}}, vol.~53, no.~3, pp. 1530--1542, 2022.

\bibitem{peng2021multiobjective}
C.~Peng and W.~Zhang, ``Multiobjective dynamic optimization of cooperative
  difference games in infinite horizon,'' \emph{\JournalTitle{IEEE Transactions
  on Systems, Man, and Cybernetics: Systems}}, vol.~51, no.~11, pp. 6669--6680,
  2021.

\bibitem{yu2021dynamic}
K.~Yu, J.~Liang, B.~Qu, Y.~Luo, and C.~Yue, ``Dynamic selection
  preference-assisted constrained multiobjective differential evolution,''
  \emph{\JournalTitle{IEEE Transactions on Systems, Man, and Cybernetics:
  Systems}}, vol.~52, no.~5, pp. 2954--2965, 2021.

\bibitem{yang2022local}
S.~Yang, H.~Huang, F.~Luo, Y.~Xu, and Z.~Hao, ``Local-diversity evaluation
  assignment strategy for decomposition-based multiobjective evolutionary
  algorithm,'' \emph{\JournalTitle{IEEE Transactions on Systems, Man, and
  Cybernetics: Systems}}, vol.~53, no.~3, pp. 1697--1709, 2022.

\bibitem{Deb2007DNSGA2}
K.~Deb, S.~Karthik \emph{et~al.}, ``Dynamic multi-objective optimization and
  decision-making using modified nsga-ii: a case study on hydro-thermal power
  scheduling,'' in \emph{International Conference on Evolutionary
  Multi-criterion Optimization}.\hskip 1em plus 0.5em minus 0.4em\relax
  Springer, 2007, pp. 803--817.

\bibitem{Jiang2016SGEA}
S.~Jiang and S.~Yang, ``A steady-state and generational evolutionary algorithm
  for dynamic multiobjective optimization,'' \emph{\JournalTitle{IEEE
  Transactions on Evolutionary Computation}}, vol.~21, no.~1, pp. 65--82, 2016.

\bibitem{Ma2021feature}
X.~Ma, J.~Yang, H.~Sun, Z.~Hu, and L.~Wei, ``Feature information prediction
  algorithm for dynamic multi-objective optimization problems,''
  \emph{\JournalTitle{European Journal of Operational Research}}, vol. 295,
  no.~3, pp. 965--981, 2021.

\bibitem{Hu2023layered}
Y.~Hu, J.~Zheng, S.~Jiang, S.~Yang, and J.~Zou, ``Handling dynamic
  multiobjective optimization environments via layered prediction and
  subspace-based diversity maintenance,'' \emph{\JournalTitle{IEEE Transactions
  on Cybernetics}}, vol.~53, no.~4, pp. 2572--2585, 2023.

\bibitem{Zhang2020evo}
K.~Zhang, C.~Shen, X.~Liu, and G.~G. Yen, ``Multi-objective evolution strategy
  for dynamic multi-objective optimization,'' \emph{\JournalTitle{IEEE
  Transactions on Evolutionary Computation}}, vol.~24, no.~5, pp. 974--988,
  2020.

\bibitem{Hatzakis2006FL}
I.~Hatzakis and D.~Wallace, ``Dynamic multi-objective optimization with
  evolutionary algorithms: a forward-looking approach,'' in \emph{Proceedings
  of The 8th Annual Conference on Genetic and Evolutionary Computation}, 2006,
  pp. 1201--1208.

\bibitem{Zhou2014PPS}
A.~Zhou, Y.~Jin, and Q.~Zhang, ``A population prediction strategy for
  evolutionary dynamic multiobjective optimization,'' \emph{\JournalTitle{IEEE
  Transactions on Cybernetics}}, vol.~44, no.~1, pp. 40--53, 2014.

\bibitem{Muruganantham2016KF}
A.~Muruganantham, K.~C. Tan, and P.~Vadakkepat, ``Evolutionary dynamic
  multiobjective optimization via kalman filter prediction,''
  \emph{\JournalTitle{IEEE Transactions on Cybernetics}}, vol.~46, no.~12, pp.
  2862--2873, 2016.

\bibitem{Rambabu2019mixture}
R.~Rambabu, P.~Vadakkepat, K.~C. Tan, and M.~Jiang, ``A mixture-of-experts
  prediction framework for evolutionary dynamic multiobjective optimization,''
  \emph{\JournalTitle{IEEE Transactions on Cybernetics}}, vol.~50, no.~12, pp.
  5099--5112, 2019.

\bibitem{zhang2022solving}
Q.~Zhang, X.~He, S.~Yang, Y.~Dong, H.~Song, and S.~Jiang, ``Solving dynamic
  multi-objective problems using polynomial fitting-based prediction
  algorithm,'' \emph{\JournalTitle{Information Sciences}}, vol. 610, pp.
  868--886, 2022.

\bibitem{Cao2019MOEADSVR}
L.~Cao, L.~Xu, E.~D. Goodman, C.~Bao, and S.~Zhu, ``Evolutionary dynamic
  multiobjective optimization assisted by a support vector regression
  predictor,'' \emph{\JournalTitle{IEEE Transactions on Evolutionary
  Computation}}, vol.~24, no.~2, pp. 305--319, 2019.

\bibitem{liu2023cooperative}
X.-F. Liu, J.~Zhang, and J.~Wang, ``Cooperative differential evolution with an
  attention-based prediction strategy for dynamic multiobjective
  optimization,'' \emph{\JournalTitle{IEEE Transactions on Systems, Man, and
  Cybernetics: Systems}}, 2023.

\bibitem{Rong2018multidirectional}
M.~Rong, D.~Gong, Y.~Zhang, Y.~Jin, and W.~Pedrycz, ``Multidirectional
  prediction approach for dynamic multiobjective optimization problems,''
  \emph{\JournalTitle{IEEE Transactions on Cybernetics}}, vol.~49, no.~9, pp.
  3362--3374, 2018.

\bibitem{Hu2020solving}
Y.~Hu, J.~Ou, J.~Zheng, J.~Zou, S.~Yang, and G.~Ruan, ``Solving dynamic
  multi-objective problems with an evolutionary multi-directional search
  approach,'' \emph{\JournalTitle{Knowledge-Based Systems}}, vol. 194, p.
  105175, 2020.

\bibitem{Zhang2007moead}
Q.~Zhang and H.~Li, ``Moea/d: A multiobjective evolutionary algorithm based on
  decomposition,'' \emph{\JournalTitle{IEEE Transactions on Evolutionary
  Computation}}, vol.~11, no.~6, pp. 712--731, 2007.

\bibitem{Goh2009dCOEA}
C.-K. Goh and K.~C. Tan, ``A competitive-cooperative coevolutionary paradigm
  for dynamic multiobjective optimization,'' \emph{\JournalTitle{IEEE
  Transactions on Evolutionary Computation}}, vol.~13, no.~1, pp. 103--127,
  2009.

\bibitem{zheng2023dynamic}
J.~Zheng, F.~Zhou, J.~Zou, S.~Yang, and Y.~Hu, ``A dynamic multi-objective
  optimization based on a hybrid of pivot points prediction and diversity
  strategies,'' \emph{\JournalTitle{Swarm and Evolutionary Computation}},
  vol.~78, p. 101284, 2023.

\bibitem{Liu2018pso}
R.~Liu, J.~Li, J.~Fan, and L.~Jiao, ``A dynamic multiple populations particle
  swarm optimization algorithm based on decomposition and prediction,''
  \emph{\JournalTitle{Applied Soft Computing}}, vol.~73, pp. 434--459, 2018.

\bibitem{chen2019multi}
L.~Chen, Q.~Li, X.~Zhao, Z.~Fang, F.~Peng, and J.~Wang, ``Multi-population
  coevolutionary dynamic multi-objective particle swarm optimization algorithm
  for power control based on improved crowding distance archive management in
  crns,'' \emph{\JournalTitle{Computer Communications}}, vol. 145, pp.
  146--160, 2019.

\bibitem{Gong2019similarity}
D.~Gong, B.~Xu, Y.~Zhang, Y.~Guo, and S.~Yang, ``A similarity-based cooperative
  co-evolutionary algorithm for dynamic interval multiobjective optimization
  problems,'' \emph{\JournalTitle{IEEE Transactions on Evolutionary
  Computation}}, vol.~24, no.~1, pp. 142--156, 2019.

\bibitem{Shang2014quantum}
R.~Shang, L.~Jiao, Y.~Ren, L.~Li, and L.~Wang, ``Quantum immune clonal
  coevolutionary algorithm for dynamic multiobjective optimization,''
  \emph{\JournalTitle{Soft Computing}}, vol.~18, no.~4, pp. 743--756, 2014.

\bibitem{xu2018memory}
X.~Xu, Y.~Tan, W.~Zheng, and S.~Li, ``Memory-enhanced dynamic multi-objective
  evolutionary algorithm based on lp decomposition,''
  \emph{\JournalTitle{Applied Sciences}}, vol.~8, no.~9, p. 1673, 2018.

\bibitem{Peng2015novel}
Z.~Peng, J.~Zheng, J.~Zou, and M.~Liu, ``Novel prediction and memory strategies
  for dynamic multiobjective optimization,'' \emph{\JournalTitle{Soft
  Computing}}, vol.~19, no.~9, pp. 2633--2653, 2015.

\bibitem{Wang2019SSA}
Y.~Wang, T.~Du, T.~Liu, and L.~Zhang, ``Dynamic multiobjective squirrel search
  algorithm based on decomposition with evolutionary direction prediction and
  bidirectional memory populations,'' \emph{\JournalTitle{IEEE Access}},
  vol.~7, pp. 115\,997--116\,013, 2019.

\bibitem{sahmoud2016memory}
S.~Sahmoud and H.~R. Topcuoglu, ``A memory-based nsga-ii algorithm for dynamic
  multi-objective optimization problems,'' in \emph{European Conference on the
  Applications of Evolutionary Computation}.\hskip 1em plus 0.5em minus
  0.4em\relax Springer, 2016, pp. 296--310.

\bibitem{zou2021reinforcement}
F.~Zou, G.~G. Yen, L.~Tang, and C.~Wang, ``A reinforcement learning approach
  for dynamic multi-objective optimization,'' \emph{\JournalTitle{Information
  Sciences}}, vol. 546, pp. 815--834, 2021.

\bibitem{Jiang2017transfer}
M.~Jiang, Z.~Huang, L.~Qiu, W.~Huang, and G.~G. Yen, ``Transfer learning-based
  dynamic multiobjective optimization algorithms,'' \emph{\JournalTitle{IEEE
  Transactions on Evolutionary Computation}}, vol.~22, no.~4, pp. 501--514,
  2017.

\bibitem{Jiang2020fast}
M.~Jiang, Z.~Wang, L.~Qiu, S.~Guo, X.~Gao, and K.~C. Tan, ``A fast dynamic
  evolutionary multiobjective algorithm via manifold transfer learning,''
  \emph{\JournalTitle{IEEE Transactions on Cybernetics}}, 2020.

\bibitem{yan2023inter}
L.~Yan, W.~Qi, J.~Liang, B.~Qu, K.~Yu, C.~Yue, and X.~Chai, ``Inter-individual
  correlation and dimension based dual learning for dynamic multi-objective
  optimization,'' \emph{\JournalTitle{IEEE Transactions on Evolutionary
  Computation}}, 2023.

\bibitem{Azzouz2017localsearch}
R.~Azzouz, S.~Bechikh, and L.~B. Said, ``A dynamic multi-objective evolutionary
  algorithm using a change severity-based adaptive population management
  strategy,'' \emph{\JournalTitle{Soft Computing}}, vol.~21, no.~4, pp.
  885--906, 2017.

\bibitem{Zou2017prediction}
J.~Zou, Q.~Li, S.~Yang, H.~Bai, and J.~Zheng, ``A prediction strategy based on
  center points and knee points for evolutionary dynamic multi-objective
  optimization,'' \emph{\JournalTitle{Applied Soft Computing}}, vol.~61, pp.
  806--818, 2017.

\bibitem{Li2019special}
J.~Li, R.~Liu, R.~Wang, J.~Liu, and C.~Mu, ``A special points-based hybrid
  prediction strategy for dynamic multi-objective optimization,''
  \emph{\JournalTitle{IEEE Access}}, vol.~7, pp. 62\,496--62\,510, 2019.

\bibitem{Azzouz2017dynamic}
R.~Azzouz, S.~Bechikh, and L.~B. Said, ``Dynamic multi-objective optimization
  using evolutionary algorithms: a survey,'' in \emph{Recent Advances in
  Evolutionary Multi-objective Optimization}.\hskip 1em plus 0.5em minus
  0.4em\relax Springer, 2017, pp. 31--70.

\bibitem{Sahmoud2018type}
S.~Sahmoud and H.~R. Topcuoglu, ``A type detection based dynamic
  multi-objective evolutionary algorithm,'' in \emph{International Conference
  on the Applications of Evolutionary Computation}.\hskip 1em plus 0.5em minus
  0.4em\relax Springer, 2018, pp. 879--893.

\bibitem{Ou2019novel}
J.~Ou, L.~Xing, M.~Liu, and L.~Yang, ``A novel prediction strategy based on
  change degree of decision variables for dynamic multi-objective
  optimization,'' \emph{\JournalTitle{IEEE Access}}, vol.~8, pp.
  13\,362--13\,374, 2019.

\bibitem{Zhang2019novel}
Q.~Zhang, S.~Yang, S.~Jiang, R.~Wang, and X.~Li, ``Novel prediction strategies
  for dynamic multiobjective optimization,'' \emph{\JournalTitle{IEEE
  Transactions on Evolutionary Computation}}, vol.~24, no.~2, pp. 260--274,
  2019.

\bibitem{Jiang2019SDP}
S.~Jiang, M.~Kaiser, S.~Yang, S.~Kollias, and N.~Krasnogor, ``A scalable test
  suite for continuous dynamic multiobjective optimization,''
  \emph{\JournalTitle{IEEE Transactions on Cybernetics}}, vol.~50, no.~6, pp.
  2814--2826, 2019.

\bibitem{jiang2017strength}
S.~Jiang and S.~Yang, ``A strength pareto evolutionary algorithm based on
  reference direction for multiobjective and many-objective optimization,''
  \emph{\JournalTitle{IEEE Transactions on Evolutionary Computation}}, vol.~21,
  no.~3, pp. 329--346, 2017.

\bibitem{jiang2016convergence}
------, ``Convergence versus diversity in multiobjective optimization,'' in
  \emph{International Conference on Parallel Problem Solving from
  Nature}.\hskip 1em plus 0.5em minus 0.4em\relax Springer, 2016, pp. 984--993.

\bibitem{Zhang2008rm}
Q.~Zhang, A.~Zhou, and Y.~Jin, ``Rm-meda: A regularity model-based
  multiobjective estimation of distribution algorithm,''
  \emph{\JournalTitle{IEEE Transactions on Evolutionary Computation}}, vol.~12,
  no.~1, pp. 41--63, 2008.

\bibitem{Huband2006WFG}
S.~Huband, P.~Hingston, L.~Barone, and L.~While, ``A review of multiobjective
  test problems and a scalable test problem toolkit,'' \emph{\JournalTitle{IEEE
  Transactions on Evolutionary Computation}}, vol.~10, no.~5, pp. 477--506,
  2006.

\bibitem{geraci2018measuring}
M.~V. Geraci and J.-Y. Gnabo, ``Measuring interconnectedness between financial
  institutions with bayesian time-varying vector autoregressions,''
  \emph{\JournalTitle{Journal of Financial and Quantitative Analysis}},
  vol.~53, no.~3, pp. 1371--1390, 2018.

\bibitem{kuschnig2021bvar}
N.~Kuschnig and L.~Vashold, ``Bvar: Bayesian vector autoregressions with
  hierarchical prior selection in r,'' \emph{\JournalTitle{Journal of
  Statistical Software}}, vol. 100, pp. 1--27, 2021.

\bibitem{Li2022}
\BIBentryALTinterwordspacing
J.~Li, R.~Liu, and R.~Wang, ``A change type-based self-adaptive response
  strategy for dynamic multi-objective optimization,''
  \emph{\JournalTitle{Know. Based Syst.}}, vol. 243, no.~C, may 2022. [Online].
  Available: \url{https://doi.org/10.1016/j.knosys.2022.108447}
\BIBentrySTDinterwordspacing

\bibitem{Sahmoud2019severity}
S.~Sahmoud and H.~R. Topcuoglu, ``Exploiting characterization of dynamism for
  enhancing dynamic multi-objective evolutionary algorithms,''
  \emph{\JournalTitle{Applied Soft Computing}}, vol.~85, p. 105783, 2019.

\bibitem{elgamal2015analysis}
T.~Elgamal and M.~Hefeeda, ``Analysis of pca algorithms in distributed
  environments,'' \emph{\JournalTitle{arXiv preprint arXiv:1503.05214}}, 2015.

\bibitem{korobilis2020sign}
D.~Korobilis \emph{et~al.}, \emph{Sign restrictions in high-dimensional vector
  autoregressions}.\hskip 1em plus 0.5em minus 0.4em\relax Rimini Centre for
  Economic Analysis, 2020.

\bibitem{Jiang2018DF}
S.~Jiang, S.~Yang, X.~Yao, K.~C. Tan, M.~Kaiser, and N.~Krasnogor, ``Benchmark
  functions for the cec'2018 competition on dynamic multiobjective
  optimization,'' Newcastle University, Tech. Rep., 2018.

\bibitem{Rong2020multimodel}
M.~Rong, D.~Gong, W.~Pedrycz, and L.~Wang, ``A multimodel prediction method for
  dynamic multiobjective evolutionary optimization,'' \emph{\JournalTitle{IEEE
  Transactions on Evolutionary Computation}}, vol.~24, no.~2, pp. 290--304,
  2020.

\bibitem{Jiang2016JY}
S.~Jiang and S.~Yang, ``Evolutionary dynamic multiobjective optimization:
  Benchmarks and algorithm comparisons,'' \emph{\JournalTitle{IEEE Transactions
  on Cybernetics}}, vol.~47, no.~1, pp. 198--211, 2016.

\bibitem{Deb2002NSGA2}
K.~Deb, A.~Pratap, S.~Agarwal, and T.~Meyarivan, ``A fast and elitist
  multiobjective genetic algorithm: Nsga-ii,'' \emph{\JournalTitle{IEEE
  Transactions on Evolutionary Computation}}, vol.~6, no.~2, pp. 182--197,
  2002.

\end{thebibliography}
\end{document}